\documentclass{article}

\usepackage{microtype}
\usepackage{graphicx}
\usepackage{subfigure}
\usepackage{booktabs} 

\usepackage{hyperref}
\usepackage{xcolor}         
\usepackage{multirow}
\usepackage{diagbox}
\usepackage{subfigure}
\usepackage{bbm}
\usepackage{xspace}
\usepackage{enumitem}
\usepackage{colortbl}

\usepackage[accepted]{icml2024}

\usepackage{amsmath}
\usepackage{amssymb}
\usepackage{mathtools}
\usepackage{amsthm}

\usepackage[capitalize]{cleveref}

\theoremstyle{plain}
\newtheorem{theorem}{Theorem}[section]
\newtheorem{proposition}[theorem]{Proposition}

\theoremstyle{definition}
\newtheorem{definition}[theorem]{Definition}
\newtheorem{assumption}[theorem]{Assumption}
\theoremstyle{remark}

\usepackage{amsmath,amsfonts,bm}

\def\eqref#1{equation~\ref{#1}}

\def\1{\bm{1}}

\DeclareMathAlphabet{\mathsfit}{\encodingdefault}{\sfdefault}{m}{sl}
\SetMathAlphabet{\mathsfit}{bold}{\encodingdefault}{\sfdefault}{bx}{n}

\DeclareMathOperator*{\argmax}{arg\,max}

\usepackage[textsize=tiny]{todonotes}

\newcommand{\method}{GLWS\xspace}

\newcommand{\cellc}{\cellcolor{lightgray!50}}

\icmltitlerunning{A General Framework for Learning from Weak Supervision}

\begin{document}

\twocolumn[
\icmltitle{A General Framework for Learning from Weak Supervision}




\begin{icmlauthorlist}
\icmlauthor{Hao Chen}{cmu}
\icmlauthor{Jindong Wang}{msft,wm}
\icmlauthor{Lei Feng}{sutd}
\icmlauthor{Xiang Li}{cmu}
\icmlauthor{Yidong Wang}{pku}
\\
\icmlauthor{Xing Xie}{msft}
\icmlauthor{Masashi Sugiyama}{riken,tokyo}
\icmlauthor{Rita Singh}{cmu}
\icmlauthor{Bhiksha Raj}{cmu,mbz}
\end{icmlauthorlist}

\icmlaffiliation{cmu}{Carnegie Mellon University}
\icmlaffiliation{mbz}{Mohamed bin Zayed University of AI}
\icmlaffiliation{msft}{Microsoft Research}
\icmlaffiliation{sutd}{Singapore University of Technology and Design}
\icmlaffiliation{riken}{RIKEN AIP}
\icmlaffiliation{tokyo}{The University of Tokyo}
\icmlaffiliation{pku}{Peking University}
\icmlaffiliation{wm}{William \& Mary}

\icmlcorrespondingauthor{Hao Chen}{haoc3@andrew.cmu.edu}

\icmlkeywords{Machine learning, Weakly supervised learning, Weak supervision}

\vskip 0.3in
]



\printAffiliationsAndNotice{}  

\begin{abstract}

Weakly supervised learning generally faces challenges in applicability to various scenarios with diverse weak supervision and in scalability due to the complexity of existing algorithms, thereby hindering the practical deployment.
This paper introduces a \underline{g}eneral framework for \underline{l}earning from \underline{w}eak \underline{s}upervision (\method) with a novel algorithm. 
Central to \method is an Expectation-Maximization (EM) formulation, adeptly accommodating various weak supervision sources, including instance partial labels, aggregate statistics, pairwise observations, and unlabeled data. 
We further present an advanced algorithm that significantly simplifies the EM computational demands using a Non-deterministic Finite Automaton (NFA) along with a forward-backward algorithm, which effectively reduces time complexity from quadratic or factorial often required in existing solutions to linear scale. 
The problem of learning from arbitrary weak supervision is therefore converted to the NFA modeling of them.
\method not only enhances the scalability of machine learning models but also demonstrates superior performance and versatility across 11 weak supervision scenarios. 
We hope our work paves the way for further advancements and practical deployment in this field.
Code is available at: \url{https://github.com/Hhhhhhao/General-Framework-Weak-Supervision}.

\end{abstract}

\section{Introduction}
\label{sec:intro}

Over the past few years, machine learning models have shown promising performance in virtually every aspect of our lives \cite{radford2021learning,rombach2022high,dehghani2023scaling,openai2023gpt4}.
This success is typically attributed to large-scale and high-quality training data with complete and accurate supervision. 
However, obtaining such precise labels in realistic applications is often prohibitive due to various factors, such as the cost of annotation \cite{settles2008active,gadre2023datacomp}, the biases and subjectivity of annotators \cite{tommasi2017deeper,pagano2023bias}, and privacy concerns \cite{ mireshghallah2020privacy,strobel2022data}.
The resulting \textit{incomplete}, \textit{inexact}, and \textit{inaccurate} forms of supervision are typically referred to as \textit{weak supervision} \cite{zhou2018brief,bookSugiyama+etal2022}.



\begin{figure}
    \centering
    \includegraphics[width=0.95\columnwidth]{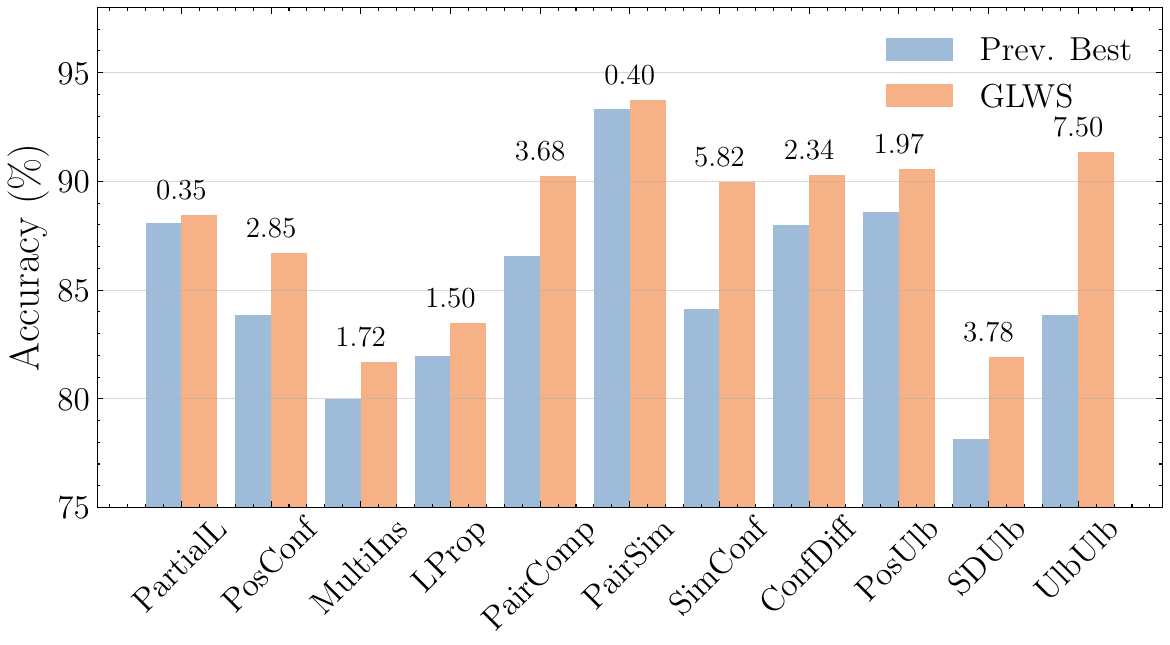}
    \vspace{-0.15in}
    \caption{Average performance overview of the proposed method on 11 common weak supervision settings, compared to previous best methods (margins shown on the top of bars). \method is capable of learning from any weak supervision universally and effectively.}
    \label{fig:teaser}
\vspace{-0.25in}
\end{figure}

Previous literature has explored numerous configurations of weak supervision problems, including learning from sets of instance label candidates \cite{Luo2010LearningFC,cour2011learning,ishida2019complementary,feng2020learning,feng2020provably,wang2022pico,revisitpllwu22l}, aggregate group statistics \cite{maron1997framework,zhou2004multi,kuck2005statistics,quadrianto2008estimating,ilse2018attention,zhang2020aggre,scott2020learning,zhang2022learning}, pairwise observations \cite{bao2018classification,bao2020pairwise,feng2021pointwise,cao2021learning,wang2023binary}, and unlabeled data \cite{lu2018minimal,sohn2020fixmatch,shimada2021classification,usb2022,tang2023multi}.
More recently, some efforts have been made to design versatile techniques that can handle multiple settings simultaneously \cite{van2018theory,zhang2020aggre,chiang2023unified,ShuklaDAE23,uumwei23a}.

Despite the prosperous developments in various settings, we identify two challenges that impede the practical application of these weakly supervised methods.
First, designing a method capable of universally handling all configurations remains difficult.
The variation in forms of weak supervision often necessitates specialized and tailored solutions \cite{ilse2018attention, yan2018deep, yang2022attention, zhang2022learning, scott2020learning}.
Even recent versatile solutions are limited in their applicability to certain contexts \cite{ShuklaDAE23, uumwei23a}. 
Second, prior works typically exhibit limited scalability in realistic problems due to oversimplifications and unfavorable modeling complexity.
Some methods assume conditional independence of instances for aggregate observations \cite{van2018theory, cui2020classification, zhang2020aggre, uumwei23a}, making them unsuitable for handling long sequence data prevalent in practical scenarios.
Moreover, despite such simplifications, they still require infeasible computational complexity, either quadratic \cite{ShuklaDAE23} or factorial \cite{uumwei23a}\footnote{We compare the complexity in terms of the naive implementation. There might be practical implementation techniques which can reduce the complexity effectively while being applicable to all the dynamic programming methods, including GLWS.}, to address specific weak supervision configurations.

\begin{figure}[t!]
    \centering
    \includegraphics[width=\columnwidth]{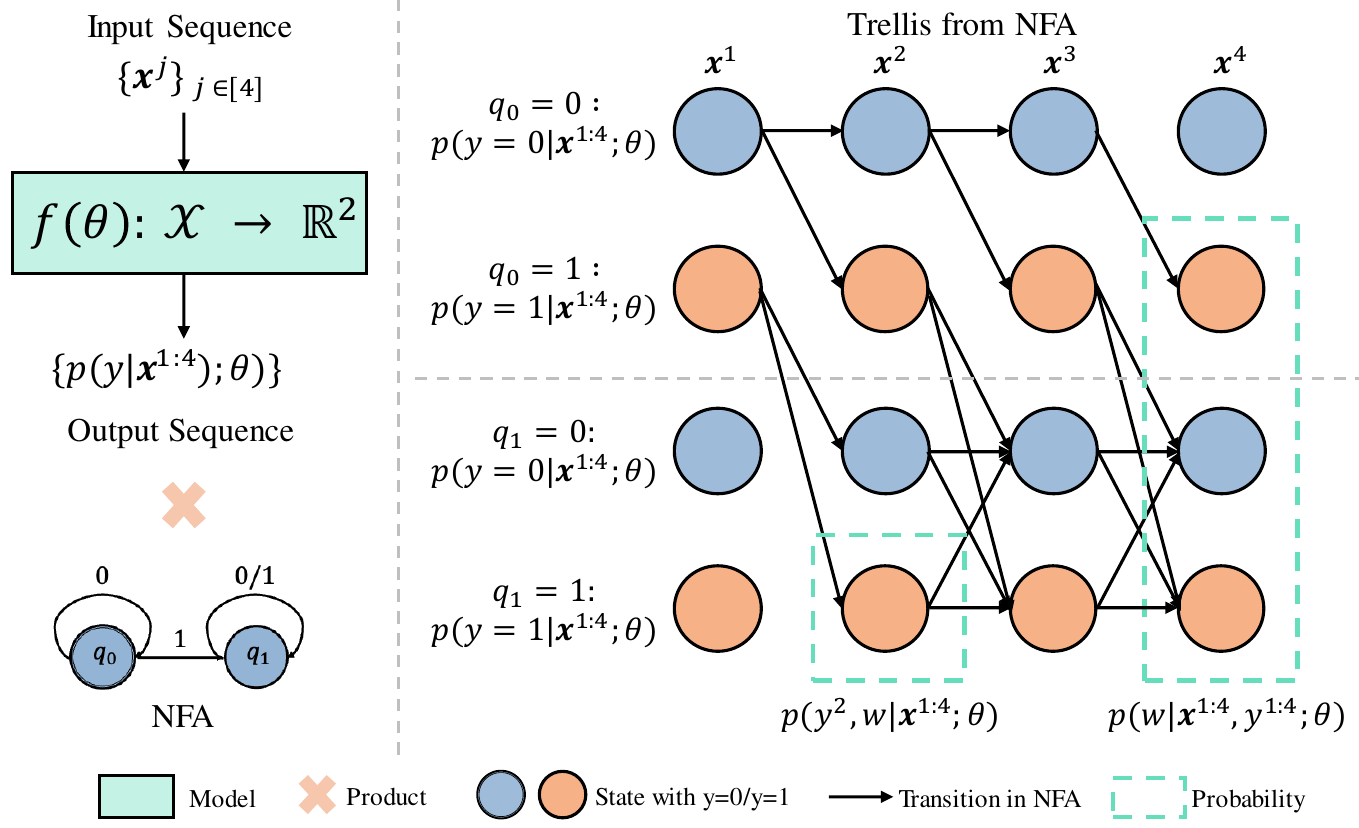}
    \vspace{-0.2in}
    \caption{Overview of \method for learning from arbitrary weak supervision. We model weak supervision as a Non-deterministic Finite Automaton (NFA). By taking the product of the prediction sequence and NFA, we can utilize the forward-backward algorithm to solve the proposed complete EM formulation in linear time.}
    \label{fig:overview}
\vspace{-0.25in}
\end{figure}

To overcome these challenges and effectively apply weakly supervised learning in real-world scenarios, we propose a \underline{g}eneral framework and a novel algorithm that allows efficient \underline{l}earning from arbitrary \underline{w}eak \underline{s}upervision, termed as \method, as in the results demonstrated in \cref{fig:teaser}.
At the core of \method is an Expectation-Maximization (EM) \cite{dempster1977maximum} learning objective formulation for weak supervision, and a forward-backward algorithm \cite{rabiner1989tutorial,graves2006ctc} designed to solve the EM in linear time by representing arbitrary form of weak supervision as a Non-deterministic Finite Automaton (NFA) \cite{rabin1959finite}.
More specifically, to train a classification model with learnable parameters $\theta$ on weak supervision, denoted abstractly as $W$, we treat the ground truth label $Y$ as a missing latent variable and maximize the log-likelihood of joint input $X$ and $W$: $\log P(X, W;\theta) = \log \sum_Y P(X, W;\theta) P(Y| X, W;\theta)$.  
As \(P(Y| X, W;\theta)\) is unknown before determining \(\theta\), solving the problem usually requires iterative hill-climbing solutions. 
Therefore, we employ the widely used EM algorithm, which iteratively maximizes the expectation of the log-likelihood \(\mathbb{E}_{Y|X,W;\theta^t}[ \log P(X, W, Y;\theta)]\) at time step $t$.
It leads to two training objectives: an unsupervised instance consistency term \(\mathbb{E}_{Y|X,W;\theta^t}[\log P(Y | X;\theta)]\) that encourages the prediction to be consistent with the labeling distribution imposed by $W$, and a supervised objective \(\log P(W | Y, X;\theta)\) that fosters the group predictions fulfilling $W$.
We further propose a novel perspective to perform the EM formulation. 
Without loss of generality, we treat both the inputs and the precise labels as sequence\footnote{For aggregated and pairwise observations, the inputs are naturally sequences of instances. The inputs can be viewed as permutation-invariant sequences at the batch (dataset) level for weak supervision of partial labels and unlabeled data. The same applies to the precise labels and predictions from the model.}. 
Thus, the problem of identifying all possible labelings is converted into assigning labels/symbols to the input sequence in a manner that adheres to $W$. 
This process can be effectively modeled using an NFA \cite{rabin1959finite}, where the finite set of states and transition is dictated by $W$, and the finite set of symbols corresponds to $Y$. 
The EM learning objectives can then be computed efficiently in linear time using a forward-backward algorithm on the \emph{trellis} expanded from the NFA and model's predictions. 
An overview is shown in \cref{fig:overview}.

While this is not the first EM perspective of weak supervision \cite{denoeux2011maximum,quost2016clustering,wang2022pico,chen2023imprecise,uumwei23a}, 
\method distinguishes from prior arts in solving the \emph{complete EM efficiently and practically}. 
Compared to the recent efforts towards the unification of weak supervision, our method neither relies on the aforementioned conditional independence assumption as in \citet{uumwei23a} nor involves approximation of EM as in \citet{wang2022pico,ShuklaDAE23} that solves the supervised term of the proposed EM only. 

Our contributions can be summarized as:
\begin{itemize}[leftmargin=1em]
\setlength\itemsep{0em}
    \item We propose \method, a unified EM framework that accommodates weak supervision of arbitrary forms, leading to two learning objectives, as a generalization of the prior arts on weak supervision. 
    \item We design a forward-backward algorithm that performs the EM by treating weak supervision as an NFA. The EM can thus be computed via iterative forward-backward pass on the trellis expanded from the NFA in linear time.
    \item On \textbf{11} weak supervision settings, the proposed method consistently achieves the state-of-the-art performance, demonstrating its universality and effectiveness. Our codebase covering all these settings is released.
\end{itemize}






\section{Related Work}
\label{sec:related}

\subsection{Learning from Weak Supervision}
Various problems for learning from weak supervision have been extensively studied in the past, and we categorize them into four broad categories: instance label candidates, aggregate observations, pairwise observations, and unlabeled data. 
Learning from instance label candidates, also known as partial label (PartialL) or complementary label (CompL) learning \cite{cour2011learning,Luo2010LearningFC,Feng2020ProvablyCP,Wang2019AdaptiveGG,wen2021leveraged,revisitpllwu22l,wang2022pico,ishida2019complementary,feng2020learning}, involves weak supervision as a set of label candidates, either containing or complementary to the ground truth label for each instance.
Aggregate observation assumes supervision over a group of instances \cite{zhang2020aggre}, with multiple instances (MultiIns) learning \cite{maron1997framework,ilse2018attention} and label proportion (LProp) learning \cite{quadrianto2008estimating,scott2020learning,zhang2022learning} as common examples.
The weak supervision here usually denotes statistics over a group of instances. 
Pairwise observation, a special case of aggregate observation, deals with pairs of instances. 
Pairwise comparison (Pcomp) \cite{feng2021pointwise} and pairwise similarity (PSim) \cite{bao2018classification,zhang2020aggre}, along with more recent developments such as similarity confidence (SimConf) \cite{cao2021learning} and confidence difference (ConfDiff) \cite{wang2023binary}, fall into this category.
Similarity, comparison, confidence scores, and relationships from the pre-trained models are usually adopted as weak supervision for pairwise observations. 
The fourth category, unlabeled data, is often supplemented by the labeled dataset as the weak supervision in this setting, which is sometimes complemented by the class's prior information.
Semi-supervised learning (SemiSL) \cite{sohn2020fixmatch,xie2020unsupervised,zhang2021flexmatch,wang2023freematch,chen2023softmatch}, positive unlabeled (PosUlb) learning \cite{du2015convex,hammoudeh2020learning,chen2020variational,garg2021mixture,kiryo2017positive,zhao2022dist}, similarity dissimilarity unlabeled (SDUlb) learning \cite{shimada2021classification}, and Unlabeled unlabeled (UlbUlb) learning \cite{lu2018minimal,tang2023multi} fall into this category.
Our framework is capable of addressing and unifying these diverse categories.

\subsection{Towards the Unification of Weak Supervision}
\label{sec:related-uni}

Although researchers have invested significant efforts in finding solutions to different forms of weak supervision, the practical unification of these problems still remains a distant goal.
PosUlb, SDUlb, and UlbUlb learning can be connected to each other by substituting parameters \cite{lu2018minimal,feng2021pointwise}. 
\citet{zhang2020aggre} have developed a probabilistic framework for pairwise \cite{hsu2019multi} and triplet comparison \cite{cui2020classification}.
\citet{ShuklaDAE23} proposed a unified solution for weak supervision involving count statistics. 
They used a dynamic programming method over the aggregate observation to compute and maximize the count loss of \(P(W|Y, X;\theta)\), corresponding to the supervised term in our EM formulation. 
The computational complexity is thus quadratic to the group length since the proposed dynamic programming algorithm iterates through the entire group.
\citet{uumwei23a} introduced the universal unbiased method (UUM) for aggregate observation, which is also interpretable from the EM perspective. 
Based on the assumptions of conditional independence of instances within a group and weak supervision given true labels, \citet{uumwei23a} derived closed-form objectives for MultiIns, LProp, and PSim settings. 
However, the oversimplification of conditional independence limits UUM's scalability, particularly for LProp learning with long sequences.
\citet{chiang2023unified} provided a comprehensive risk analysis for various types of weak supervision from the perspective of the contamination matrix.
Our framework offers a versatile and scalable solution, capable of efficiently handling a wider range of weak supervision without the limitations imposed by oversimplifications or computational complexity.

\section{Method}
\label{sec:method}

In this section, we introduce our proposed framework and algorithm for learning from arbitrary weak supervision (\method).
\method is based on the EM formulation \cite{dempster1977maximum}, where we consider the precise labels as the latent variable. 
We introduce an NFA \cite{rabin1959finite} modeling of weak supervision, which allows us to compute EM using the forward-backward algorithm in linear time.

\subsection{Preliminaries}
\label{sec:method-pre}

Let $\mathbf{x} \in \mathcal{X}$ be a training instance and $y \in \mathcal{Y}$ the corresponding precise supervision, where the input space $\mathcal{X} \subset \mathbb{R}^D$ has $D$ dimensions, and the label space $\mathcal{Y} = [K-1] := \{0, 1, \dots, K-1\}$ encompasses a total of $K$ classes. 
In fully supervised learning, the training dataset with complete and precise annotations is defined as $\mathcal{D} = \{(\mathbf{x}_i, y_i)\}_{i \in [N]}$ and consists of $N$ samples. 
Assume that each training example $(\mathbf{x}, y)$ is identically and independently sampled from the joint distribution $p(\mathbf{x}, y)$.
The classifier $f(\theta): \mathcal{X} \rightarrow \mathbb{R}^K$ predicts $p(y|\mathbf{x};\theta)$ with learnable parameters $\theta$, and is trained to maximize the log-likelihood $\log P(X, Y;\theta)$:
\begin{equation}
    \theta^* = \argmax_{\theta} \log P(X, Y;\theta).
\vspace{-0.1in}
\end{equation}
This process results in the cross-entropy (CE) loss function:
\begin{equation}
  \mathcal{L}_{\mathrm{Full}} =  \sum_{i=1}^N \sum_{k=0}^K   - \mathbbm{1}[y_i = k] \log p(y_i |\mathbf{x}_i;\theta).
\vspace{-0.1in}
\end{equation}

\begin{figure*}[t!]
    \centering
    
    \hfill
    \subfigure[Partial label]{\label{fig:nfa_partial}\includegraphics[width=0.24\textwidth]{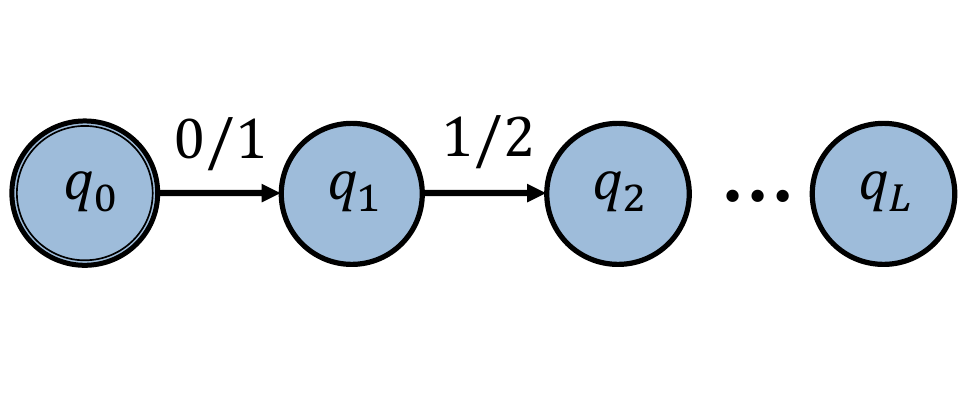}}
    \hfill
    \subfigure[Multiple instance]{\label{fig:nfa_multi_ins}\includegraphics[width=0.24\textwidth]{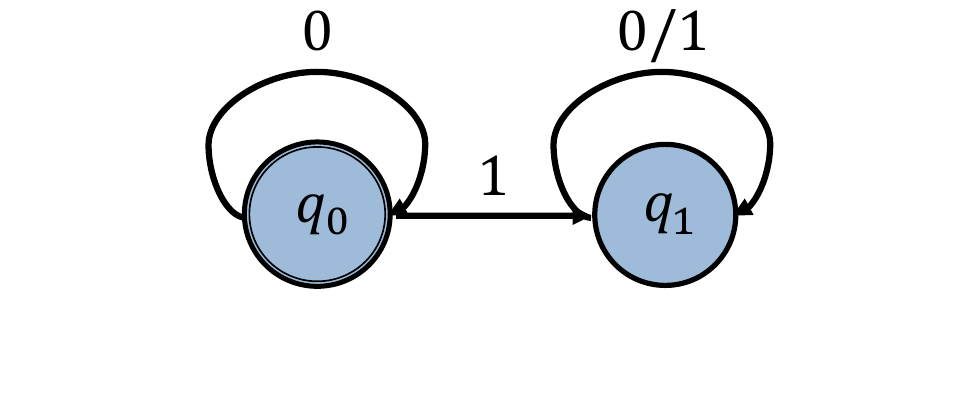}}
    \hfill
    \subfigure[Label proportion]{\label{fig:nfa_proportion}\includegraphics[width=0.24\textwidth]{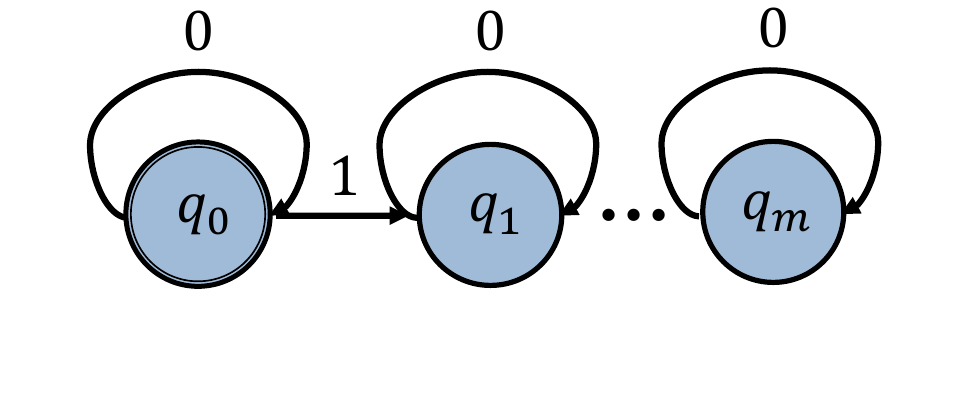}}
    \hfill
    \subfigure[Pairwise comp.]{\label{fig:nfa_pcomp}\includegraphics[width=0.24\textwidth]{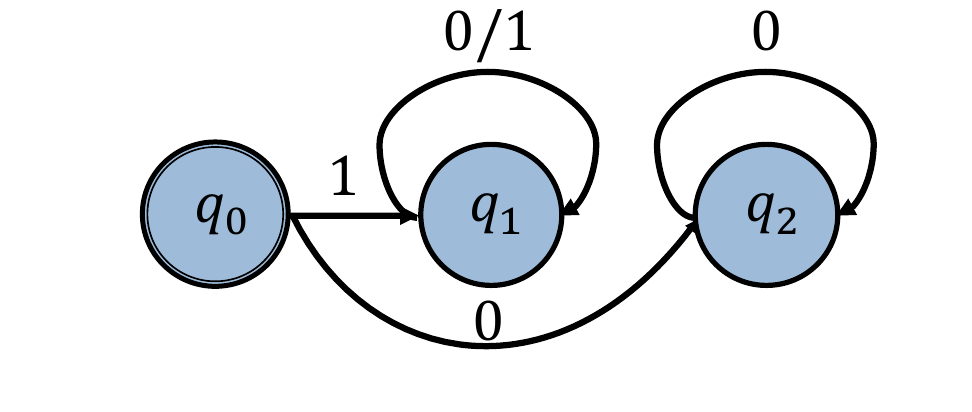}}
    \hfill
\\
    \hfill
    \subfigure[Pairwise sim. (w/ conf. $c$)]{\label{fig:nfa_psim}\includegraphics[width=0.24\textwidth]{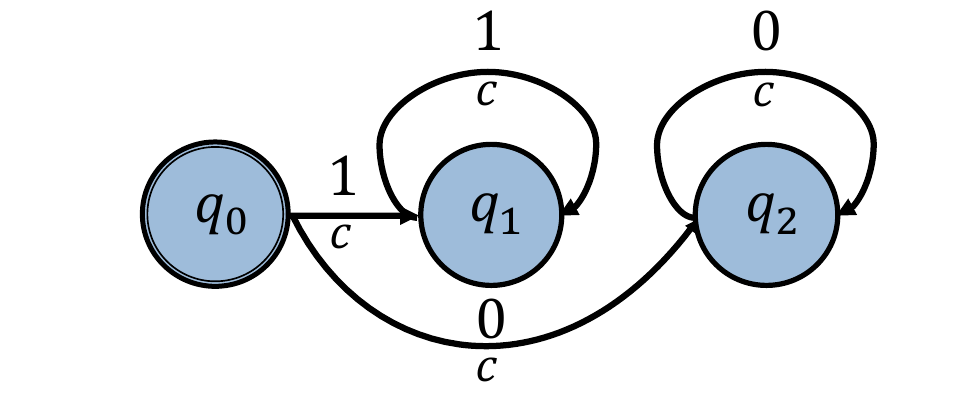}}
    \hfill
    \subfigure[Pairwise dissim. (w/ conf. $c$)]{\label{fig:nfa_pdsim}\includegraphics[width=0.24\textwidth]{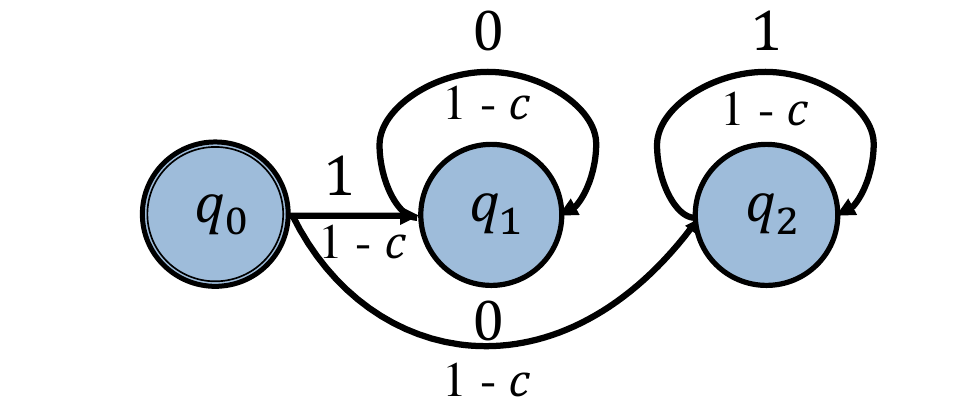}}
    \hfill
    \subfigure[Positive conf.  (w/ conf. $c$)]{\label{fig:nfa_pos_conf}\includegraphics[width=0.24\textwidth]{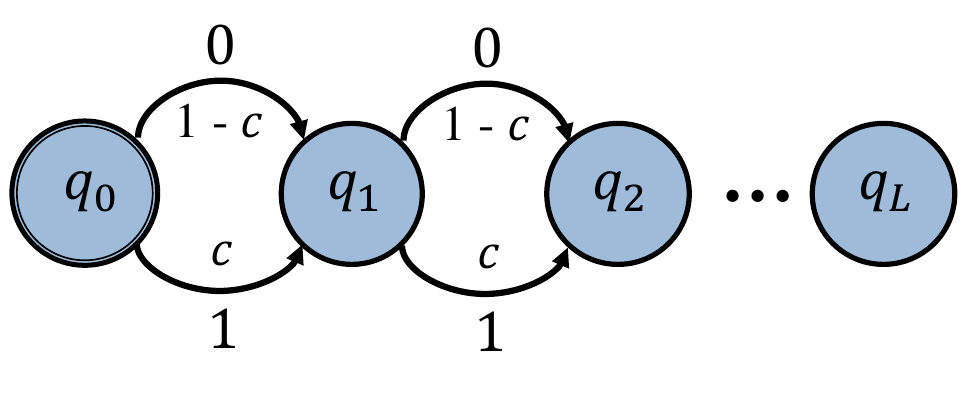}}
    \hfill
    \subfigure[Unlabeled data (w/ prior $p$)]{\label{fig:nfa_pu}\includegraphics[width=0.24\textwidth]{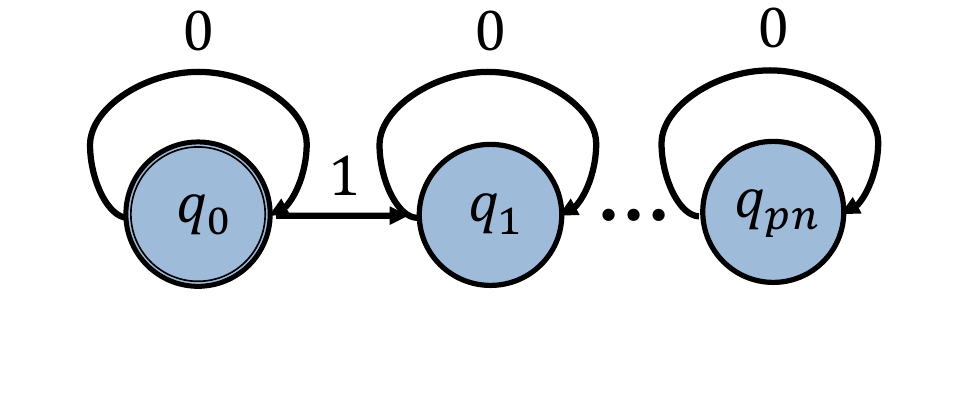}}
    \hfill
    \vspace{-0.1in}
    \caption{NFA for common weak supervision types for a sequence input of size $L$. (a) Partial labels, where the NFA has $L$ transitions for each input with partial labels as symbols; (b) Multiple instances, whose NFA has 2 states, and can only transit to the accepting state via $1$ to ensure at least one positive instance in the sequence; (c) Label proportion, whose NFA has $m+1$ states for $m$ positive samples in the sequence; (d) Pairwise comparison, whose NFA has 3 states and covers $\{(1, 1), (1, 0), (0, 0)\}$; (e) Pairwise similarity with confidence score $c$. The NFA also has 3 states and covers $\{(1, 1), (0, 0)\}$. If $c$ is given as in similarity confidence and confidence difference, each edge is weighted by $c$; (f) Pairwise dissimilarity with confidence $c$ for $\{(1, 0), (0, 1)\}$; (g) Positive confidence, whose NFA also has $L$ transitions weighted by confidence $c$; (h) Unlabeled data with class prior $p$. The NFA is equivalent to expectation of label count as $pn$. 
    }
    \label{fig:nfa}
\vspace{-0.2in}
\end{figure*}

\subsection{General Framework for Weak Supervision}
\label{sec:method-em}

In practice, we may not have fully accessible precise supervision, i.e., $Y$ is unknown. 
Instead, we may encounter various types of weak supervision for training instances, e.g., instance-wise label candidates, aggregated count statistics, pairwise similarity, unlabeled data, etc. 
We define weak supervision abstractly as $W$, representing an arbitrary form of information given to the training instances. 
For example, in PartialL \cite{feng2020provably,wang2022pico,revisitpllwu22l}, $W$ is given as a set of label candidates for each instance $\mathcal{S} \subseteq \mathcal{Y} $.
In MultiIns \cite{maron1997framework,zhou2004multi} and LProp learning \cite{yu2014learning} that deals with aggregate observations, $W$ is given as the count statistics for each label $\{\sum_{j=1}^L \mathbbm{1}[y^j = k] \ge 1 \mid \forall k \in \mathcal{Y} \}$ and $\{ \sum_{j=1}^L \mathbbm{1}[y^j = k] \mid \forall k \in \mathcal{Y} \}$ over a group of $L$ instances $\{\mathbf{x}^j\}_{j \in [L]}$\footnote{We use $\mathbf{x}_i$ ($y_i$) to denote an instance/group in dataset of size $N$, and $\mathbf{x}^j$ ($y^j$) to denote an instance in the group $\{\mathbf{x}^j\}_{j \in [L]}$ of size $L$. Each $\mathbf{x}_i$ ($y_i$) in the dataset can denote a group with $L \ge 1$.}, respectively. 
When $W$ represents the precise labels, it recovers fully supervised learning. 
With $W$, we must estimate the model to maximize the likelihood of the data $X$ and the information $W$ we have been provided:
\begin{equation}
\begin{split}
    \theta^* &= \argmax_\theta \log P(X, W;\theta) \\
             &= \argmax_\theta \sum\nolimits_Y \log P(X, W, Y; \theta).
\end{split}
\label{eq:weaksup}
\end{equation}
As $Y$ is unknown and the marginalization over $Y$ requires $\theta$, it is infeasible to solve \cref{eq:weaksup} in a closed form, and instead typically needs the iterative hill-climbing solutions like EM algorithm.
Thus, the maximum log-likelihood estimation in \cref{eq:weaksup} can be solved by iteratively maximizing the variational lower bound of the log-likelihood $\log P(X, W, Y;\theta)$:
\begin{equation}
    \theta^{t+1} = \argmax_{\theta} \mathbb{E}_{Y|X, W;\theta^t} \left[ \log P(X, W, Y; \theta) \right],
\label{eq:em}
\end{equation}
where $\theta^t$ denotes the $t$-th estimation of $\theta$. 
$P(Y|X, W;\theta^t)$ represents a distribution on all possible labelings imposed by $W$ with $\theta^t$.
The log-likelihood is then maximized on the expectation over the distribution of all possible labelings.
The derivation of \cref{eq:em} is provided in \cref{sec:appendix-proof-theorm-em}. 

To derive the loss function for arbitrary weak supervision that includes instance-level and group-level $W$ from \cref{eq:em}, without loss of generality, we treat the realization of training instances $X$ and the corresponding true precise labels $Y$ all as sequence: $\mathbf{x}^{1:L} = \{\mathbf{x}^j\}_{j \in [L]}$ and $y^{1:L} = \{y^j\}_{j \in [L]}$ of size $L$ and $L$ could be 1. 
We also treat different types of weak supervision $w \in W$ as the information given for the input sequence.
The sequence can be naturally formed from a batch of training samples, an aggregate observation, or a pairwise observation. 
Each instance in the dataset is thus generalized to $\mathbf{x}_i \rightarrow \mathbf{x}_i^{1:L} $ with $L \geq 1$.
We make the following assumption, which almost always holds in reality:
\begin{assumption}
\label{assump}
    The sequence of predictions on precise labels $y^{1:L}$ is conditionally independent given the whole sequence of inputs $\mathbf{x}^{1:L}$, i.e., $p(y^{1:L} | \mathbf{x}^{1:L}) = \prod_{j}^L p(y^j | \mathbf{x}^{1:L})$.
\vspace{-0.2in}
\end{assumption}
Note that this assumption is accurate if computed by any model where the predictions made for any instance are not fed back to the model as input when computing other predictions. 
This notation allows us to deal with different weak supervision for both instance and group data more flexibly. 
\begin{proposition}
For weakly supervised learning problems, the training objectives can be derived from \cref{eq:em} as:
\begin{equation}
\begin{split}
    & \mathcal{L}_{\mathrm{Weak}} = \mathcal{L}_{\mathrm{U}} + \mathcal{L}_{\mathrm{S}}, \\
     & \mathcal{L}_{\mathrm{U}} = \sum_{i=1}^N \sum_{j=1}^L -p(y_i^j| \mathbf{x}^{1:L}_i, w_i;\theta^t) \log p(y_i^j | \mathbf{x}^j_i;\theta),  \\
    & \mathcal{L}_{\mathrm{S}} = \sum_i^N -\log p( w_i | \mathbf{x}_{i}^{1:L}, y_i^{1:L};\theta). 
\end{split}
\label{eq:em-loss}
\end{equation}
\label{proposition:em-loss}
\vspace{-0.2in}
\end{proposition}
The detailed derivation of \cref{eq:em-loss} is shown in \cref{sec:appendix-proof-proposition-em-loss}. 
\cref{eq:em-loss} consists of two parts: an unsupervised loss $\mathcal{L}_{\mathrm{U}}$ that encourages the instance-wise predictions from the classifier to align with the probability of this prediction given all possible labelings imposed by $W$, and a supervised loss $\mathcal{L}_{\mathrm{S}}$ that encourages the sequence predictions to fulfill $W$.

\subsection{Weak Supervision as NFA}
\label{sec:method-nfa}

Although the proposed EM formulation can deal with various types of weak supervision flexibly, it is still computational intensive to calculate the probability $p(y^j | \mathbf{x}^{1:L}, w;\theta)$ and $p(w |y^{1:L}, \mathbf{x}^{1:L};\theta)$ for all possible labelings imposed by the given weak supervision. 
For example, in LProp learning where $W$ is the label count over a group of $L$ instances, the complexity of finding all possible labelings is of factorial $\mathcal{O}(L!)$. 
In most cases, the complexity is of exponential $\mathcal{O}(K^L)$ where $K$ is the total number of classes. 
Moreover, while some recent methods towards unification can also be related to the proposed EM formulation \cite{ShuklaDAE23,uumwei23a}, they both involve a certain degree of simplification to approximate the complete EM formulation, which limits their scalability, as discussed in \cref{sec:related-uni}.
Our method notably distinguishes from the prior arts in that we tackle weak supervision with the \textit{complete} EM.


Here, we present a novel perspective to overcome the infeasibility of computing the complete EM. 
Under the sequential view, we treat the problem of assigning labels $\{y^j\}_{j \in [L]}$ to inputs $\{\mathbf{x}^j\}_{j \in [L]}$ as generating a sequence of symbols $y^{1:L}$ to $\mathbf{x}^{1:L}$ fulfilling $W$. 
For simplicity, we only consider binary classification problems here. 
In \cref{sec:method-class}, we will show the generalization to multi-class classification problems. 
This process naturally fits the mechanism of the NFA \cite{rabin1959finite}. 
We can thus model weak supervision $W$ as an NFA that defines a set of finite states and transition rules, summarizing all possible labelings imposed by $W$.


\begin{definition}
\cite{rabin1959finite}
A Non-deterministic Finite Automaton (NFA) is defined as a tuple ($Q$, $\Sigma$, $\delta$, $q_0$, $F$), where $Q$ is a finite set of states, $\Sigma$ is a finite set of symbols, $\delta$ is a transition function $Q \times \Sigma \rightarrow P(Q)$, $q_0 \in Q$ is the initial state, and $F \subseteq Q$ is a set of accepting states.
\end{definition}

We define the NFA of weak supervision $W$ similarly, with states $Q$, initial state $q_0$, and accepting states $F$ determined by $W$, symbols $\Sigma = \mathcal{Y} = \{0,1\}$, and a transition function $\delta$ defining the possible transitions between states.
We can now represent all possible labelings imposed by $W$ as the language accepted by the NFA: $\{y^{1:L} | \delta(q_0, y^{1:L}) \in F \}$.
The problem of finding all possible labelings is thereby converted to modeling the NFA of different types of weak supervision.
We present the modeling of NFA for common forms of $W$ in \cref{fig:nfa}.
For example, in MultiIns learning (\cref{fig:nfa_multi_ins}) with $W$ denoting at least one positive sample within a group instance, its NFA contains 2 states $Q=\{q_0, q_1\}$.
The initial state $q_0$ can only transit to the accepting state $q_1$ via symbol $1$ to ensure $W$ is satisfied. 
Once reaching $q_1$, transit via $0$ and $1$ are both allowed. 
For LProp with $m$ positive labels (\cref{fig:nfa_proportion}), its NFA must transit via 1 for $m$ times from $q_0$ to $q_{m}$ to satisfy $W$, resulting in $m+1$ states.

\subsection{The Forward-Backward Algorithm}
\label{sec:method-fb}

We are now set to compute the EM formulation with NFA. 
\begin{proposition}
Given the inputs $\mathbf{x}^{1:L}$, we treat the outputs sequence from the classifier $p(y^{1:L}|\mathbf{x}^{1:L};\theta)$ as a linear chain graph. 
By taking the product on the linear chain graph of $p(y^{1:L}|\mathbf{x}^{1:L};\theta)$ and the NFA graph of $W$, we obtain the trellis in the resulting graph as possible labelings. 
\end{proposition}

We have $p(y^j | \mathbf{x}^{1:L}, w;\theta^t) \propto p(y^j , w | \mathbf{x}^{1:L};\theta^t)$ from Bayes' theorem, where the latter denotes the total probability of all valid labelings that go through $y^j$, and $p(w | y^{1:L}, \mathbf{x}^{1:L};\theta)$ in \cref{eq:em-loss} denotes the total probability from accepting states of the resulting graph. 
Fortunately, both the probability of $p(y^j , w | \mathbf{x}^{1:L};\theta^t)$ and $p(w | y^{1:L}, \mathbf{x}^{1:L};\theta)$ can be computed in linear time to the sequence length with dynamic programming on the trellis of the resulting graph, specifically the forward-backward algorithm \cite{rabiner1989tutorial,graves2006ctc}. 
The core idea of the forward-backward algorithm is that the sum over paths corresponding to a labeling can be broken down into iterative sum over paths corresponding to the prefixes and the postfixes of that labeling. 
Thus, the probabilities can be obtained iteratively in linear time. 

\begin{figure}[t!]
    \centering
    \includegraphics[width=\columnwidth]{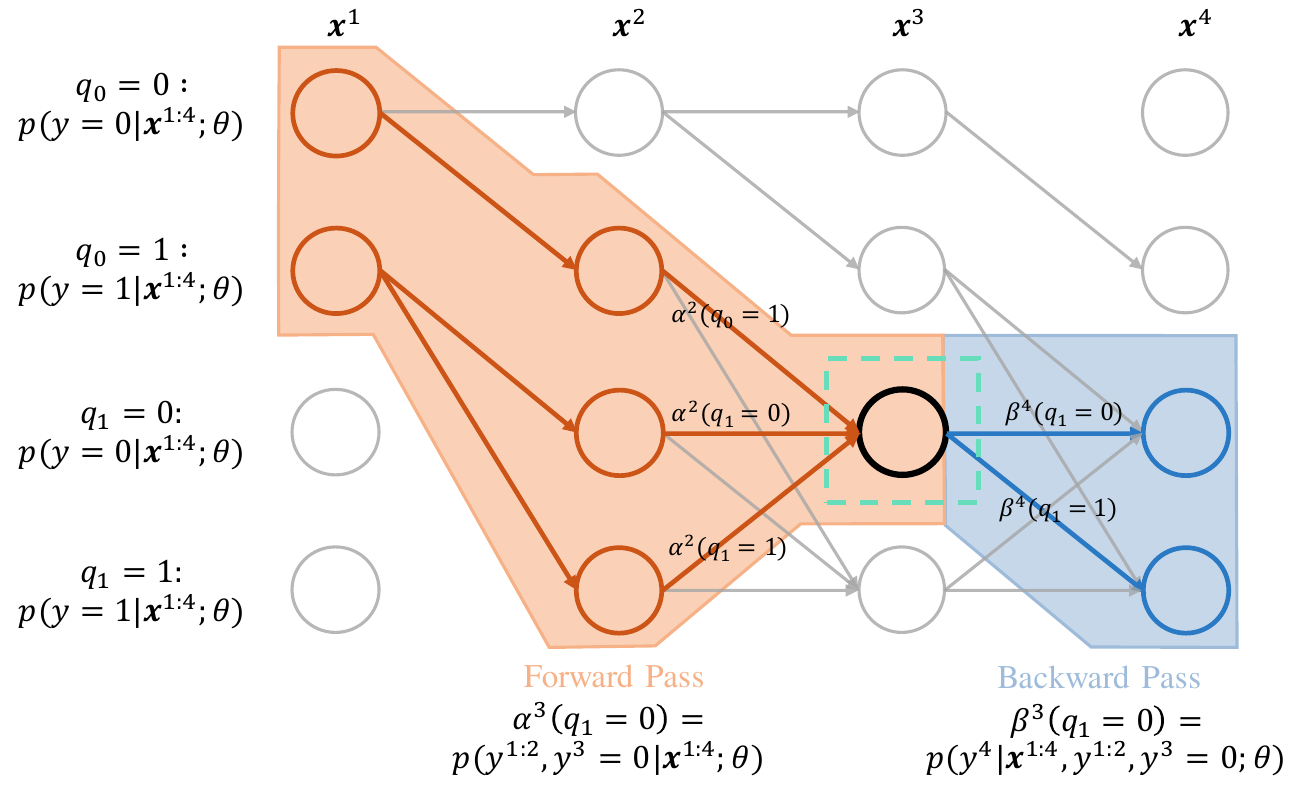}
    \vspace{-0.1in}
    \caption{Illustration of the forward pass and backward pass in forward-backward algorithm to compute $p(y^j, w | \mathbf{x}^{1:L};\theta^t)$.}
    \label{fig:fb}
\vspace{-0.1in}
\end{figure}

We illustrate the \textit{trellis} expanded from the NFA of $W$ in MultiIns learning with $L=4$, with the help of \cref{fig:overview} (more illustrations on other settings are shown in \cref{sec:appendix-method-trellis}), and the process of the forward-backward algorithm as shown in \cref{fig:fb}.
The resulting graph has 4 states at each step of the sequence, where the first two correspond to $q_0$, the others to $q_1$, and the trellis to the transition rules in NFA. 
Each path from $\mathbf{x}^1$ to $\mathbf{x}^L$ denotes an available labeling. 
To compute the probabilities, we define the forward score $\alpha^j(q=y)$\footnote{Here, $q = y$ is a shorthanded notation for $\mathbf{x}^j$ transiting to the next state via $y$.} and backward score $\beta^j(q=y)$ for each state $q$ at step $j$:
\begin{equation}
\resizebox{.98\columnwidth}{!}{$
\begin{split}
\alpha^j(q=y) &= \sum_{ y' \in \{ y^{i-1} | \delta(q_0, y^{1:L}) \in F \}}  \alpha^{i-1}(q=y')  p(y^j | \mathbf{x}^{1:L}; \theta^t), \\
\hat{\beta}^j(q=y) &= \sum_{ y' \in \{ y^{j+1} | \delta(q_0, y^{1:L}) \in F \} }  \hat{\beta}^{j + 1}(q=y') p(y^j | \mathbf{x}^{1:L}; \theta^t), \\
\beta^j(q=y) &= \frac{\hat{\beta}^j(q=y)}{p(y^j | \mathbf{x}^{1:L}; \theta^t)},
\end{split}
$}
\end{equation}
where $\hat{\beta}^j(q=y)$ is used as a proxy for easier computation of $\beta^j(q=y)$ \cite{graves2006ctc}. 
The forward score $\alpha^{j}(q=y)$ indicates the total probability of all preceding labeling that fulfills $W$ at $j$-th inputs with $p(y^{1:j} | \mathbf{x}^{1:L}; \theta^t)$, and correspondingly, the backward score $\beta^{j}(q=y)$ indicates the total probability of all succeeding labeling that fulfills $W$ at $j$-th inputs given the preceding $p(y^{j+1:L} | \mathbf{x}^{1:L}, y^{1:j};\theta^t)$, $\forall y^j \in \{ y^j | \delta(q_0, y^{1:L}) \in F \}$.
Both $\alpha^j(q=y)$ and $\beta^j(q=y)$ can be calculated recursively through the forward and backward pass on the graph, with \textit{linear complexity of $\mathcal{O}(|Q| L)$}, where $|Q|$ is the number of states on the NFA of $W$. 
The joint probability at each position of the sequence thus can be calculated as:
\begin{equation}
\resizebox{.9\columnwidth}{!}{$
p(y^j, w | \mathbf{x}^{1:L};\theta^t) = \sum_{q \in Q}  \frac{ \alpha^{j}(q=y^j)\beta^{j} (q=y^j) }{\sum_{y' \in \mathcal{Y}   }   \alpha^{j}(q=y')\beta^{j} (q=y') }.
$}
\end{equation}
Moreover, the probability for supervised objective can also be easily computed as the summation of the probabilities at the accepting nodes on the graph with linear complexity:
\begin{equation}
    p(w | y^{1:L}, \mathbf{x}^{1:L}; \theta) = \sum_{q \in F}  \sum_{y' \in \mathcal{Y}} \alpha^{L}(q=y').
\end{equation}
Now, we can bring these quantities back to \cref{eq:em-loss} to perform training. 
In practice, we implement the forward-backward algorithm in log space and adopt the re-scaling strategy \cite{mcauley2013hidden} for numerical stability.
We present the pseudo-algorithm of the forward-backward process of the common settings in \cref{sec:appendix-method-alg}.

\subsection{Extension to Multi-Class or Multi-Label Scenarios}
\label{sec:method-class}

In the analysis above, we model the NFA of $W$ only for binary classification problems. 
Here, we demonstrate how to extend the modeling to multi-class (multi-label) classification problems. 
While it is natural to extend to multiple classes, for example, for partial labels as shown in \cref{fig:nfa_partial}, it is not straightforward to directly model the NFA of $W$ with more than two values in its symbols $\Sigma$ for aggregate observations.
Considering the example with MultiIns learning, where the group of instances has two multi-class labels: at least one cat and at least one dog, the complexity of $|Q|$ in the NFA modeling will increase exponentially, thus also increasing the complexity in computing the loss functions. 
Instead, to deal with it, we treat each class as a separate positive class and other classes as a negative class, build an NFA on this class, and train each class as a binary classification problem with binary cross-entropy (BCE) loss. 
This is the common technique widely adopted in pre-training \cite{wightman2110resnet,touvron2022deit} and we demonstrate its effectiveness in \cref{sec:exp-aggregate} for weakly supervised learning.

\section{Experiments}
\label{sec:experiments}

In this section, we demonstrate the universality and effectiveness of the proposed method comprehensively on various weakly supervised learning settings. 
We conduct the evaluation mainly on CIFAR-10 \cite{krizhevsky2009learning}, CIFAR-100 \cite{krizhevsky2009learning}, STL-10 \cite{coates2011analysis}, and ImageNet-100 \cite{russakovsky2015imagenet}. 
Results on MNIST \cite{deng2012mnist} and F-MNIST \cite{xiao2017online} are included in the Appendix, where most of the baseline methods were evaluated. 
We compare our method (\method) on $11$ weak supervision settings of partial labels in \cref{sec:exp-partial}, aggregate observations in \cref{sec:exp-aggregate}, pairwise observations in \cref{sec:exp-pairwise}, and unlabeled data in \cref{sec:exp-unlabeled}. 
Additionally, we provide more analysis and discussion in \cref{sec:exp-analysis}. 
We develop a codebase
for implementations and experiments of all baselines and the proposed method, which will be open-sourced. 
Experiments are conducted three times with the average performance and standard deviation reported.

\begin{table}[t!]
\centering
\caption{Accuracy on partial label (PartialL) learning for instance-wise weak supervision. All results are averaged over three runs.}
\label{tab:partial-label}
\resizebox{\columnwidth}{!}{%
\begin{tabular}{@{}l|cc|cc|cc|cc@{}}
\toprule
Dataset &
  \multicolumn{2}{c|}{CIFAR-10} &
  \multicolumn{2}{c|}{CIFAR-100} &
  \multicolumn{2}{c|}{STL-10} &
  \multicolumn{2}{c}{ImageNet-100} \\ \midrule
Ratio &
   0.50 &
   0.70 &
   0.10 &
   0.20 &
   0.10 &
   0.30 &
   0.01 &
   0.05 \\ \midrule
CC &
   92.51\scriptsize{$\pm$0.04} &
   89.01\scriptsize{$\pm$0.20} &
   77.44\scriptsize{$\pm$0.32} &
   74.60\scriptsize{$\pm$0.17} &
   77.02\scriptsize{$\pm$0.69} &
   73.26\scriptsize{$\pm$0.34} &
   73.14\scriptsize{$\pm$0.94} &
   64.67\scriptsize{$\pm$0.74} \\
LWS &
   85.66\scriptsize{$\pm$0.32} &
   80.71\scriptsize{$\pm$0.10} &
   50.67\scriptsize{$\pm$0.33} &
   43.51\scriptsize{$\pm$0.32} &
   67.65\scriptsize{$\pm$0.33} &
   58.18\scriptsize{$\pm$1.65} &
   72.04\scriptsize{$\pm$0.77} &
   62.13\scriptsize{$\pm$0.95} \\
PRODEN &
   93.32\scriptsize{$\pm$0.23} &
   90.26\scriptsize{$\pm$0.20} &
   77.50\scriptsize{$\pm$0.15} &
   74.89\scriptsize{$\pm$0.13} &
   77.44\scriptsize{$\pm$0.26} &
   73.19\scriptsize{$\pm$1.05} &
   78.61\scriptsize{$\pm$0.63} &
   77.59\scriptsize{$\pm$0.60} \\
PiCO &
   93.85\scriptsize{$\pm$0.60} &
   91.11\scriptsize{$\pm$0.70} &
   77.80\scriptsize{$\pm$0.31} &
   74.99\scriptsize{$\pm$0.57} &
   77.74\scriptsize{$\pm$0.52} &
   74.18\scriptsize{$\pm$0.41} &
   80.93\scriptsize{$\pm$0.81} &
   78.74\scriptsize{$\pm$1.34} \\
RCR &
   94.04\scriptsize{$\pm$0.02} &
   91.45\scriptsize{$\pm$0.10} &
   78.03\scriptsize{$\pm$0.07} &
   75.40\scriptsize{$\pm$0.12} &
   78.02\scriptsize{$\pm$0.40} &
   74.67\scriptsize{$\pm$0.56} &
   81.52\scriptsize{$\pm$0.94} &
   79.67\scriptsize{$\pm$1.22} \\
\cellc \method &
\cellc   \textbf{94.31\scriptsize{$\pm$0.09}} &
\cellc   \textbf{92.06\scriptsize{$\pm$0.14}} &
\cellc   \textbf{78.35\scriptsize{$\pm$0.11}} &
\cellc   \textbf{75.82\scriptsize{$\pm$0.25}} &
\cellc   \textbf{78.56\scriptsize{$\pm$0.27}} &
\cellc   \textbf{74.79\scriptsize{$\pm$0.21}} &
\cellc   \textbf{82.66\scriptsize{$\pm$0.54}} &
\cellc   \textbf{81.09\scriptsize{$\pm$0.50}} \\
\bottomrule
\end{tabular}%
}
\vspace{-0.2in}
\end{table}

\subsection{Partial Labels}
\label{sec:exp-partial}

\textbf{Setup}. 
Here, we evaluate the proposed method of PartialL learning for multi-class classification, where $W$ is a set of label candidates for each training instance. 
Following \citet{revisitpllwu22l} and \citet{lv2020progressive}, we generate synthetic uniform partial labels for each dataset. We uniformly select labels other than the ground truth label with a specified partial ratio. For baselines, we adopt CC \cite{Feng2020ProvablyCP}, LWS \cite{wen2021leveraged}, PRODEN \cite{lv2020progressive}, PiCO \cite{wang2022pico}, and RCR \cite{revisitpllwu22l}. We follow the hyper-parameters from \citet{revisitpllwu22l} for training all methods, with more details provided in \cref{sec:appendix-exp-partial-setup}.

\textbf{Results}. 
The main results are shown in \cref{tab:partial-label}. 
Due to space limitations, more results are presented in \cref{tab:appendix-partial} of \cref{sec:appendix-exp-partial-results}. 
Our method generally outperforms the baselines across different partial ratios, especially on the more practical ImageNet-100 with an improvement margin over RCR of \textbf{1.28}\%. 
The complete EM formulation serves as a generalized method of the prior arts.
Moreover, our method is simple and straightforward to implement, requiring no additional loss functions like the contrastive loss in PiCO or training tricks like multiple augmentations in RCR.

\subsection{Aggregate Observations}
\label{sec:exp-aggregate}

\textbf{Setup}. 
For aggregate observations, we evaluate two common settings: MultiIns learning and LProp learning. 
MultiIns learning considers $W$ as the indicator of at least one positive sample for a class in a bag of instances, while LProp learning views $W$ as the exact count or proportion of positive samples for a class within the bag. 
We form training bags with instances sampled randomly, where the bag size is Gaussian-distributed with specified parameters. 
Previous methods typically focus on binary classification in these settings. However, in our main paper, we extend this to multi-class classification (additional binary classification results are in \cref{sec:appendix-exp-aggre-results}), with $W$ being multi-labeled.
For instance, in MultiIns learning, the weak supervision could indicate that at least one positive instance for both dog and cat classes are present in a group. 
Baselines for our evaluation include Count Loss \cite{ShuklaDAE23} and UUM \cite{uumwei23a}. 
In LProp learning, we also compare against LLP-VAT \cite{tsai2020learning}. 
Details of training hyper-parameters are shown in \cref{sec:appendix-exp-aggre-setup}.

\begin{table}[t!]
\centering
\caption{Accuracy on multi-class multi-label aggregate observations of multiple instance (MultiIns) learning and label proportion (LProp) learning. All results are averaged over three runs. }
\label{tab:aggregate}
\resizebox{\columnwidth}{!}{%
\begin{tabular}{@{}l|cc|cc|cc|cc@{}}
\toprule
  Dataset &
  \multicolumn{2}{c|}{CIFAR-10} &
  \multicolumn{2}{c|}{CIFAR-100} &
  \multicolumn{2}{c|}{STL-10} &
  \multicolumn{2}{c}{ImageNet-100} \\ \cmidrule(l){1-9} 
  Dist &
  $\mathcal{N}(10,2)$ &
  \multicolumn{1}{c|}{$\mathcal{N}(20,5)$} &
  $\mathcal{N}(5,1)$ &
  $\mathcal{N}(10,2)$ &
  $\mathcal{N}(5,1)$ &
  $\mathcal{N}(10,2)$ &
  $\mathcal{N}(3,1)$ &
  $\mathcal{N}(5,1)$ \\ 
  \# Bags &
  5,000 &
  \multicolumn{1}{c|}{2,500} &
  10,000 &
  5,000 &
  2,000 &
  1,000 &
  20,000 &
  20,000 \\ \midrule

\midrule 
\multicolumn{9}{c}{Multiple Instance Learning} \\ \midrule
  Count Loss &
  86.84\scriptsize{$\pm$0.34} &
  65.97\scriptsize{$\pm$0.94} &
  52.04\scriptsize{$\pm$1.49} &
  30.66\scriptsize{$\pm$0.68} &
  73.79\scriptsize{$\pm$1.51} &
  63.80\scriptsize{$\pm$1.66} &
  71.48\scriptsize{$\pm$1.61} &
  70.58\scriptsize{$\pm$1.14} \\
  UUM &
  13.86\scriptsize{$\pm$1.31} &
  13.21\scriptsize{$\pm$0.52} &
  1.27\scriptsize{$\pm$0.29} &
  1.01\scriptsize{$\pm$0.20} &
  18.25\scriptsize{$\pm$2.58} &
  15.45\scriptsize{$\pm$1.66} &
  1.33\scriptsize{$\pm$0.17} &
  1.25\scriptsize{$\pm$0.18} \\
\cellc  \method &
\cellc  \textbf{87.15\scriptsize{$\pm$0.32}} &
\cellc  \textbf{71.88\scriptsize{$\pm$0.55}} &
\cellc  \textbf{56.28\scriptsize{$\pm$1.16}} &
\cellc  \textbf{52.29\scriptsize{$\pm$2.93}} &
\cellc  \textbf{74.66\scriptsize{$\pm$1.64}} &
\cellc  \textbf{64.35\scriptsize{$\pm$0.52}} &
\cellc  \textbf{73.92\scriptsize{$\pm$1.38}} &
\cellc  \textbf{73.08\scriptsize{$\pm$1.76}} \\ \midrule

\midrule 
\multicolumn{9}{c}{Label Proportion Learning} \\ \midrule
  LLP-VAT &
  85.33\scriptsize{$\pm$0.44} &
  \multicolumn{1}{c|}{79.70\scriptsize{$\pm$0.48}} &
  51.95\scriptsize{$\pm$2.74} &
  52.26\scriptsize{$\pm$0.46} &
  74.76\scriptsize{$\pm$0.08} &
  70.76\scriptsize{$\pm$0.78} &
  59.97\scriptsize{$\pm$3.45} &
  68.45\scriptsize{$\pm$1.82} \\
  Count Loss &
  89.46\scriptsize{$\pm$0.24} &
  \multicolumn{1}{c|}{84.54\scriptsize{$\pm$0.39}} &
  54.13\scriptsize{$\pm$1.43} &
  36.21\scriptsize{$\pm$0.49} &
  76.60\scriptsize{$\pm$0.13} &
  73.36\scriptsize{$\pm$0.33} &
  72.17\scriptsize{$\pm$0.47} &
  72.21\scriptsize{$\pm$0.91} \\
  UUM &
  - &
  \multicolumn{1}{c|}{-} &
  53.25\scriptsize{$\pm$1.96} &
  - &
  77.26\scriptsize{$\pm$0.67} &
  - &
  71.51\scriptsize{$\pm$0.94} &
  71.14\scriptsize{$\pm$1.31} \\
\cellc  \method &
\cellc  \textbf{89.77\scriptsize{$\pm$0.45}} &
\cellc  \textbf{86.41\scriptsize{$\pm$0.11}} &
\cellc  \textbf{58.25\scriptsize{$\pm$0.61}} &
\cellc  \textbf{57.14\scriptsize{$\pm$1.71}} &
\cellc  \textbf{78.27\scriptsize{$\pm$0.77}} &
\cellc  \textbf{73.70\scriptsize{$\pm$0.19}} &
\cellc  \textbf{73.93\scriptsize{$\pm$0.33}} &
\cellc  \textbf{73.09\scriptsize{$\pm$0.84}} \\
  \bottomrule
\end{tabular}%
}
\vspace{-0.2in}
\end{table}

\textbf{Results}. 
The results are presented in \cref{tab:aggregate}. 
Our method demonstrates a significant performance gain compared to baselines across various setups. 
In MultiIns learning, our method surpasses Count Loss by \textbf{1.46}\% on CIFAR-10, \textbf{12.93}\% on CIFAR-100, 0.71\% on STL-10, and \textbf{2.47}\% on ImageNet-100, showcasing its effectiveness in more complex datasets with a larger number of classes and training group sizes.
For LProp learning, it notably outperforms previous methods, with improvements of \textbf{4.50}\% on CIFAR-100 and \textbf{2.19}\% on ImageNet-100.
The oversimplified modeling of UUM, while adequate for smaller bags and datasets (e.g., sizes 3 and 5, MNIST and Fashion-MNIST as shown in \cref{tab:appendix-aggregate}), makes it struggle with larger datasets and bag sizes as shown in \cref{tab:aggregate}.
Furthermore, for bags with an average size greater than 5, LProp learning becomes computationally infeasible in UUM due to the factorial complexity.
Compared to UUM's factorial complexity and Count Loss's quadratic complexity, our proposed method efficiently addresses various settings with linear complexity.

\begin{table}[t!]
\centering
\caption{Accuracy on binary classification of pairwise comparison (PComp) and pairwise similarity (PSim) averaged over three runs.}
\label{tab:pairwise-comp-sim}
\resizebox{\columnwidth}{!}{%
\begin{tabular}{@{}l|cc|cc|cc@{}}
\toprule
Dataset &
  \multicolumn{2}{c|}{CIFAR-10} &
  \multicolumn{2}{c|}{CIFAR-100} &
  \multicolumn{2}{c}{STL-10} \\ \midrule
  
\midrule
\multicolumn{7}{c}{Pairwise Comparison} \\ \midrule
\#Pairs &
  \multicolumn{2}{c|}{20,000} &
  \multicolumn{2}{c|}{20.000} &
  \multicolumn{2}{c}{5,000} \\ 
Prior &
  0.5 &
  0.8 &
  0.5 &
  0.8 &
  0.5 &
  0.8 \\ \midrule
PComp ABS &
  91.78\scriptsize{$\pm$0.10} &
  87.37\scriptsize{$\pm$1.89} &
  81.67\scriptsize{$\pm$0.24} &
  66.06\scriptsize{$\pm$1.19} &
  79.07\scriptsize{$\pm$0.40} &
  56.45\scriptsize{$\pm$1.86} \\
PComp ReLU &
  92.18\scriptsize{$\pm$0.22} &
  90.57\scriptsize{$\pm$0.21} &
  81.77\scriptsize{$\pm$0.59} &
  66.57\scriptsize{$\pm$1.27} &
  79.68\scriptsize{$\pm$0.75} &
  67.01\scriptsize{$\pm$1.71} \\
PComp Teacher &
  93.33\scriptsize{$\pm$0.38} &
  91.35\scriptsize{$\pm$0.27} &
  78.59\scriptsize{$\pm$0.60} &
  67.43\scriptsize{$\pm$3.09} &
  77.33\scriptsize{$\pm$0.14} &
  72.88\scriptsize{$\pm$0.15} \\
PComp Unbiased &
  91.71\scriptsize{$\pm$0.48} &
  88.22\scriptsize{$\pm$0.58} &
  67.80\scriptsize{$\pm$0.07} &
  60.86\scriptsize{$\pm$2.19} &
  77.46\scriptsize{$\pm$0.19} &
  71.60\scriptsize{$\pm$0.95} \\
Rank Pruning &
  93.98\scriptsize{$\pm$0.40} &
  91.97\scriptsize{$\pm$0.27} &
  78.90\scriptsize{$\pm$0.48} &
  71.51\scriptsize{$\pm$0.73} &
  77.89\scriptsize{$\pm$0.42} &
  73.62\scriptsize{$\pm$1.38} \\
\cellc \method &
\cellc  \textbf{94.15\scriptsize{$\pm$0.10}} &
\cellc  \textbf{93.28\scriptsize{$\pm$0.38}} &
\cellc  \textbf{83.15\scriptsize{$\pm$0.16}} &
\cellc  \textbf{80.50\scriptsize{$\pm$0.20}} &
\cellc  \textbf{81.26\scriptsize{$\pm$0.54}} &
\cellc  \textbf{79.24\scriptsize{$\pm$0.87}} \\ \midrule

\midrule
\multicolumn{7}{c}{Pairwise Similarity} \\ \midrule
\#Pairs &
  \multicolumn{2}{c|}{25,000} &
  \multicolumn{2}{c|}{25.000} &
  \multicolumn{2}{c}{5,000} \\ 
Prior &
  0.4 &
  0.6 &
  0.4 &
  0.6 &
  0.4 &
  0.6 \\ \midrule
RiskSD &
  85.78\scriptsize{$\pm$1.70} &
  85.61\scriptsize{$\pm$1.34} &
  70.41\scriptsize{$\pm$0.21} &
  64.26\scriptsize{$\pm$3.81} &
  74.15\scriptsize{$\pm$3.27} &
  69.35\scriptsize{$\pm$0.32} \\
UUM &
  97.24\scriptsize{$\pm$0.23} &
  97.16\scriptsize{$\pm$0.24} &
  87.13\scriptsize{$\pm$0.40} &
  85.19\scriptsize{$\pm$2.45} &
  83.55\scriptsize{$\pm$0.80} &
  83.64\scriptsize{$\pm$0.25} \\
\cellc \method &
\cellc  \textbf{97.44\scriptsize{$\pm$0.07}} &
\cellc  \textbf{97.18\scriptsize{$\pm$0.22}} &
\cellc  \textbf{87.25\scriptsize{$\pm$0.16}} &
\cellc  \textbf{86.96\scriptsize{$\pm$0.33}} &
\cellc  \textbf{84.81\scriptsize{$\pm$0.60}} &
\cellc  \textbf{85.19\scriptsize{$\pm$0.26}} \\ \bottomrule

\end{tabular}%
}
\vspace{-0.25in}
\end{table}

\subsection{Pairwise Observations}
\label{sec:exp-pairwise}

\textbf{Setup}. 
We conduct evaluation on four common settings of pairwise observations $(\mathbf{x}^1, \mathbf{x}^2)$ for binary classification: PComp \cite{feng2021pointwise}, PSim \cite{uumwei23a}, SimConf \cite{cao2021learning}, and ConfDiff learning \cite{wang2023binary}. 
We treat a subset of classes of each dataset as the positive class, and others as the negative class. 
Details on the class split are shown in \cref{sec:appendix-data}. 
We first set a class prior, and then sample data to form the training pairs accordingly for each setting, following the baselines \cite{feng2021pointwise,uumwei23a,cao2021learning,wang2023binary}.
For PComp, $W$ indicates the unlabeled pairs that $\mathbf{x}^1$ can only be more positive than $\mathbf{x}^2$.
We adopt PComp (and its variants) \cite{feng2021pointwise} and Rank Pruning \cite{northcutt2017learning} as baselines. 
For PSim, $W$ indicates whether the instances in the pair have similar labels
or dissimilar labels.
We use RiskSD \cite{shimada2021classification} and UUM \cite{uumwei23a} as baselines for this setting.
For SimConf and ConfDiff, $W$ is the confidence score of similarity and difference between $\mathbf{x}^1$ and $\mathbf{x}^2$, respectively. 
The confidence score is given by a pre-trained model, and we follow the previous method \cite{cao2021learning,wang2023binary} to train a model on excluded data first to compute the confidence score. 
We additionally adopt CLIP \cite{radford2021learning,cherti2023reproducible} with its zero-shot confidence score. 
Since only a non-identifiable classifiers can be learned from pairwise observations, we use clustering algorithms of Hungarian matching \cite{crouse2016implementing} similar to \citet{uumwei23a} on the predictions to evaluate. 
We present more training details of these settings in \cref{sec:appendix-exp-pair-setup}.

\begin{table}[t!]
\centering
\caption{Accuracy on binary classification of similarity confidence (SimConf) and confidence difference (ConfDiff) over three runs.}
\label{tab:pairwise-conf}
\resizebox{\columnwidth}{!}{%
\begin{tabular}{@{}l|cc|cc|cc@{}}
\toprule
Dataset &
  \multicolumn{2}{c|}{CIFAR-10} &
  \multicolumn{2}{c|}{CIFAR-100} &
  \multicolumn{2}{c}{STL-10} \\ \midrule
\#Pairs &
  \multicolumn{2}{c|}{25,000} &
  \multicolumn{2}{c|}{25.000} &
  \multicolumn{2}{c}{5,000} \\ 
Prior &
  0.4 &
  0.4 &
  0.4 &
  0.4 &
  0.4 &
  0.4 \\ 
Conf Model &
  \multicolumn{1}{c}{WRN-28-2} &
  \multicolumn{1}{c|}{CLIP ViT-B-16} &
  \multicolumn{1}{c}{ResNet-18} &
  \multicolumn{1}{c|}{CLIP ViT-B-16} &
  \multicolumn{1}{c}{ResNet-18} &
  \multicolumn{1}{c}{CLIP ViT-B-16} \\ \midrule

\midrule
\multicolumn{7}{c}{Similarity Confidence} \\ \midrule
Sconf Abs &
  87.36\scriptsize{$\pm$1.22} &
  90.16\scriptsize{$\pm$1.32} &
  75.79\scriptsize{$\pm$0.27} &
  69.51\scriptsize{$\pm$0.44} &
  76.84\scriptsize{$\pm$0.75} &
  74.44\scriptsize{$\pm$0.78} \\
Sconf ReLU &
  88.56\scriptsize{$\pm$0.57} &
  90.50\scriptsize{$\pm$0.44} &
  74.95\scriptsize{$\pm$0.55} &
  69.67\scriptsize{$\pm$1.51} &
  77.40\scriptsize{$\pm$0.31} &
  75.26\scriptsize{$\pm$0.66} \\
Sconf NN Abs &
  89.04\scriptsize{$\pm$0.88} &
  89.05\scriptsize{$\pm$2.11} &
  74.55\scriptsize{$\pm$0.23} &
  68.93\scriptsize{$\pm$2.00} &
  77.55\scriptsize{$\pm$0.31} &
  75.66\scriptsize{$\pm$0.51}  \\
Sconf Unbiased &
  88.72\scriptsize{$\pm$0.52} &
  88.71\scriptsize{$\pm$0.59} &
  72.87\scriptsize{$\pm$1.30} &
  69.55\scriptsize{$\pm$0.31} &
  77.76\scriptsize{$\pm$0.40} &
  74.36\scriptsize{$\pm$0.60}  \\
\cellc \method &
\cellc  \textbf{95.97\scriptsize{$\pm$0.11}} &
\cellc  \textbf{97.88\scriptsize{$\pm$0.11}} &
\cellc  \textbf{85.58\scriptsize{$\pm$0.88}} &
\cellc  \textbf{87.94\scriptsize{$\pm$0.34}} &
\cellc  \textbf{78.64\scriptsize{$\pm$0.16}} &
\cellc  \textbf{79.06\scriptsize{$\pm$0.05}}  \\ \midrule

\midrule
\multicolumn{7}{c}{Confidence Difference} \\ \midrule

ConfDiff Abs &
  90.12\scriptsize{$\pm$4.19} &
  88.61\scriptsize{$\pm$7.50} &
  82.89\scriptsize{$\pm$0.32} &
  81.45\scriptsize{$\pm$0.26} &
  73.17\scriptsize{$\pm$2.06} &
  77.33\scriptsize{$\pm$0.74}  \\
ConfDiff ReLU &
  90.36\scriptsize{$\pm$4.07} &
  88.78\scriptsize{$\pm$7.91} &
  83.13\scriptsize{$\pm$0.27} &
  81.68\scriptsize{$\pm$0.46} &
  72.39\scriptsize{$\pm$3.06} &
  77.59\scriptsize{$\pm$0.17} \\
ConfDiff Unbiased &
  90.05\scriptsize{$\pm$5.23} &
  87.91\scriptsize{$\pm$9.03} &
  83.65\scriptsize{$\pm$0.11} &
  81.94\scriptsize{$\pm$0.43} &
  72.13\scriptsize{$\pm$2.70} &
  77.98\scriptsize{$\pm$0.08} \\
\cellc \method &
\cellc  \textbf{95.36\scriptsize{$\pm$0.19}} &
\cellc  \textbf{96.14\scriptsize{$\pm$0.67}} &
\cellc  \textbf{86.12\scriptsize{$\pm$0.76}} &
\cellc  \textbf{83.42\scriptsize{$\pm$1.12}} &
\cellc  \textbf{77.99\scriptsize{$\pm$0.75}} &
\cellc  \textbf{78.49\scriptsize{$\pm$0.31}}  \\ \bottomrule

\end{tabular}%
}
\vspace{-0.2in}
\end{table}

\textbf{Results}. 
We present the main results for PComp and Psim in \cref{tab:pairwise-comp-sim}, and for SimConf and ConfDiff in \cref{tab:pairwise-conf}.
The proposed method presents consistent and superior performance, where the improvement margin is significant especially on larger datasets.
On CIFAR-100, our method improves the previous best by \textbf{10.23}\% on pairwise comparison and by \textbf{14.03}\% on similarity confidence. 
All the baseline methods here require the class prior in the proposed loss functions, which must be given or estimated. 
Ours does not require class prior and still achieves the best performance.
More results of pairwise observations are in \cref{sec:appendix-exp-pair-results}.

\subsection{Unlabeled Data}
\label{sec:exp-unlabeled}

\textbf{Setup}. For unlabeled data, we consider the settings of binary classification where only the class prior is given to the unlabeled data as weak supervision: PosUlb \cite{du2015convex}, UlbUlb \cite{lu2018minimal}, and SDUlb learning \cite{shimada2021classification}.
We present only the results of PosUlb learning in the main paper, and other settings are shown in \cref{sec:appendix-exp-ulb-results}. 
We similarly split the classes into either the positive subset or the negative subset as pairwise observations.
For PosUlb learning, we first randomly select a specified number of positive samples as a labeled set, and treat the remaining data as an unlabeled set. 
For STL-10, we additionally add its split of extra data to the unlabeled set.
We consider Count Loss \cite{ShuklaDAE23}, CVIR \cite{garg2021mixture}, DistPU \cite{zhao2022dist}, NNPU \cite{kiryo2017positive}, UPU \cite{kiryo2017positive}, and VarPU \cite{chen2020variational} as baselines. 
More details are in \cref{sec:appendix-exp-ulb-setup}.

\begin{table}[t!]
\centering
\caption{Accuracy on positive unlabeled (PosUlb) learning for binary classification. All results are averaged over three runs.}
\label{tab:pu}
\resizebox{\columnwidth}{!}{%
\begin{tabular}{@{}l|cc|cc|cc@{}}
\toprule
 &
  \multicolumn{2}{c|}{CIFAR-10} &
  \multicolumn{2}{c|}{CIFAR-100} &
  \multicolumn{2}{c}{STL-10} \\ \midrule
\# Pos &
  500 &
  1000 &
  1000 &
  2000 &
  500 &
  1000 \\ \midrule
Count Loss &
  87.76\scriptsize{$\pm$0.59} &
  88.61\scriptsize{$\pm$0.68} &
  70.57\scriptsize{$\pm$1.50} &
  78.13\scriptsize{$\pm$0.19} &
  77.11\scriptsize{$\pm$0.60} &
  78.79\scriptsize{$\pm$0.96}  \\
CVIR &
  88.65\scriptsize{$\pm$2.59} &
  93.37\scriptsize{$\pm$0.24} &
  78.56\scriptsize{$\pm$0.22} &
  82.94\scriptsize{$\pm$0.37} &
  77.67\scriptsize{$\pm$1.11} &
  81.84\scriptsize{$\pm$1.10}  \\
Dist PU &
  83.61\scriptsize{$\pm$4.52} &
  82.60\scriptsize{$\pm$2.48} &
  69.12\scriptsize{$\pm$1.39} &
  69.83\scriptsize{$\pm$1.43} &
  71.07\scriptsize{$\pm$1.12} &
  70.89\scriptsize{$\pm$0.63} \\
NN PU &
  87.45\scriptsize{$\pm$0.66} &
  90.32\scriptsize{$\pm$0.50} &
  75.49\scriptsize{$\pm$0.88} &
  77.26\scriptsize{$\pm$0.47} &
  74.57\scriptsize{$\pm$0.54} &
  77.32\scriptsize{$\pm$0.95}  \\
U PU &
  81.38\scriptsize{$\pm$2.17} &
  87.51\scriptsize{$\pm$0.24} &
  68.70\scriptsize{$\pm$0.79} &
  70.08\scriptsize{$\pm$0.98} &
  73.37\scriptsize{$\pm$0.57} &
  75.31\scriptsize{$\pm$0.74}  \\
Var PU &
  77.00\scriptsize{$\pm$2.82} &
  84.45\scriptsize{$\pm$2.58} &
  61.02\scriptsize{$\pm$0.22} &
  66.02\scriptsize{$\pm$0.29} &
  60.98\scriptsize{$\pm$0.78} &
  62.37\scriptsize{$\pm$1.44}  \\
\cellc \method &
\cellc  \textbf{91.67\scriptsize{$\pm$0.19}} &
\cellc  \textbf{93.69\scriptsize{$\pm$0.28}} &
\cellc  \textbf{80.30\scriptsize{$\pm$0.12}} &
\cellc  \textbf{83.32\scriptsize{$\pm$0.23}} &
\cellc  \textbf{79.60\scriptsize{$\pm$0.95}} &
\cellc  \textbf{82.87\scriptsize{$\pm$0.83}}  \\ \bottomrule
\end{tabular}%
}
\vspace{-0.2in}
\end{table}

\textbf{Results}. 
On weak supervision with unlabeled data, our method also presents superior performance, as shown in \cref{tab:pu}.
Notably, our method outperforms the previous best by \textbf{3.02}\% on CIFAR-10 with 500 positive labeled data and \textbf{4.81}\% on CIFAR-100 with 1000 positive labeled data. 
Compared to Count Loss, which computes only the supervised objective in the proposed EM formulation with quadratic complexity, its performance often falls short of other baselines such as CVIR. 
Our method only requires linear time.

\subsection{Analysis and Discussion}
\label{sec:exp-analysis}


\begin{figure}[t!]
    \centering

    \hfill
    \subfigure[CIFAR-10, $\mathcal{N}(20, 5)$]{\label{fig:converge_multi_ins_cifar10}\includegraphics[width=0.49\columnwidth]{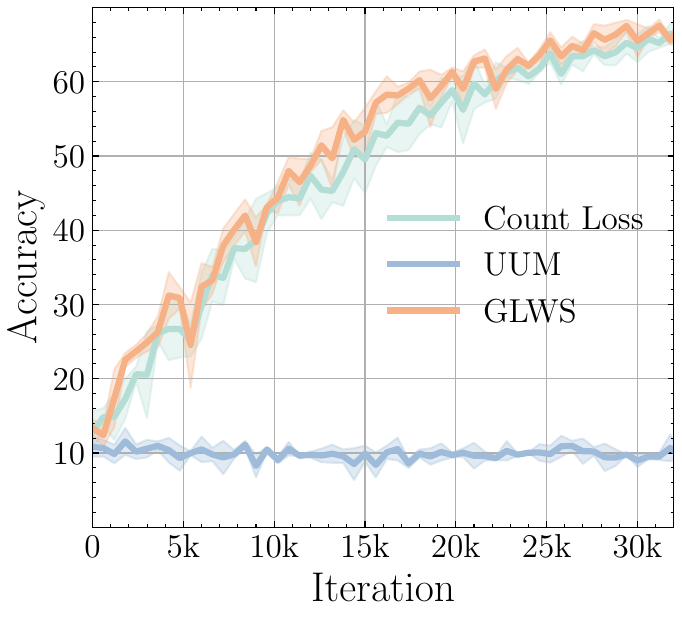}}
    \hfill
    \subfigure[CIFAR-100, $\mathcal{N}(10, 2)$]{\label{fig:converge_multi_ins_cifar100}\includegraphics[width=0.49\columnwidth]{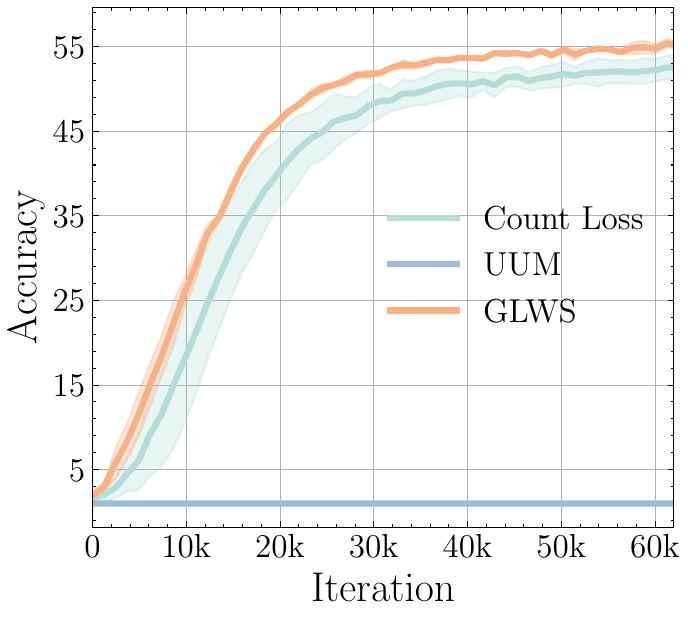}}
    \hfill
    \vspace{-0.1in}
    \caption{Convergence of accuracy with error bar on multiple instance learning with long input sequence. (a) CIFAR-10 with bag length distribution of $\mathcal{N}(20, 5)$; (b) CIFAR-100 with $\mathcal{N}(10, 2)$. Our method shows superior convergence with more stable training.}
    \label{fig:converge}
\vspace{-0.1in}
\end{figure}

\textbf{Convergence}. 
EM algorithm might be notoriously known for difficulty in convergence and converging to local minima. 
We present the convergence plots, especially for aggregate observations with long sequence lengths, to show that this is not a limitation for \method in weakly supervised learning. 
As shown in \cref{fig:converge}, our method converges faster to a better solution with a more stable training process (narrower error bars), compared to Count Loss \cite{ShuklaDAE23}.

\begin{figure}[t!]
    \centering

    \hfill
    \subfigure[Multiple instance]{\label{fig:runtime_multi_ins}\includegraphics[width=0.49\columnwidth]{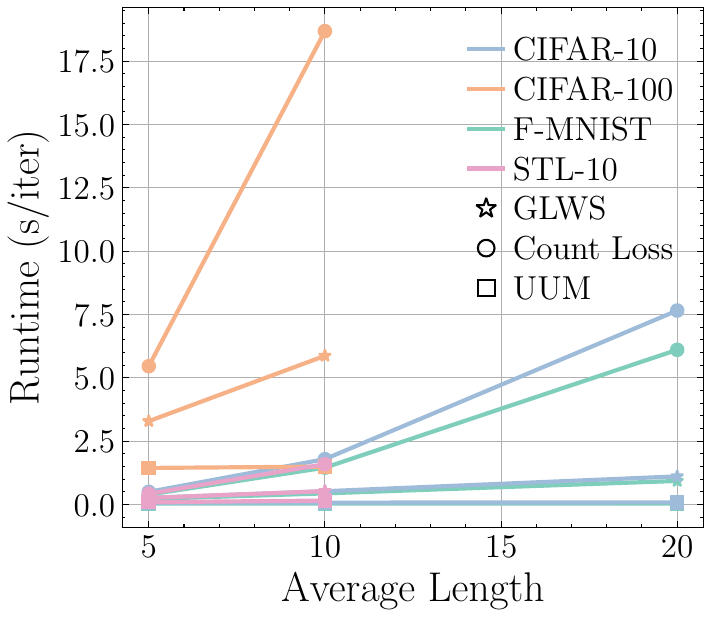}}
    \hfill
    \subfigure[Label proportion]{\label{fig:runtime_proportion}\includegraphics[width=0.49\columnwidth]{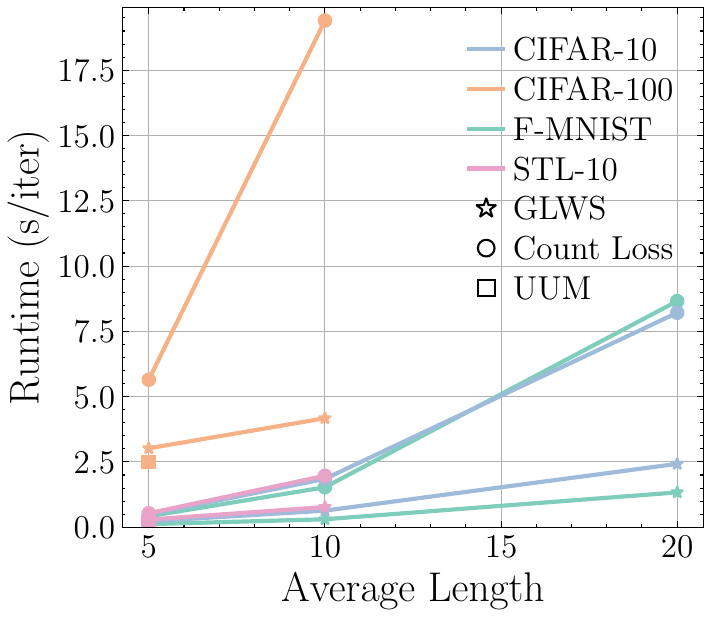}}
    \hfill
    \vspace{-0.1in}
    \caption{Runtime (s/iter.) vs. average input length for aggregate observations on evaluated datasets. (a) Multiple instance; (b) Label proportion. Our method shows a reasonable runtime trade-off. 
    }
    \label{fig:runtime}
\vspace{-0.1in}
\end{figure}

\textbf{Runtime}. 
We compare the running time explicitly in \cref{fig:runtime} for aggregate observations.
It is obvious that Count Loss \cite{ShuklaDAE23} presents (approximately) a quadratic trend in runtime as input length increases. 
UUM \cite{uumwei23a} shows consistent runtime for MultiIns learning with its oversimplification, leading to a practical performance gap as shown in \cref{tab:aggregate} and \cref{tab:appendix-aggregate}. 
On label proportion, it is only applicable to input length of 5 because of its factorial complexity. 
Ours achieves the most reasonable performance and runtime trade-off with the proposed efficient algorithm.

\textbf{Extension}. Our framework is flexibly extensible to other settings (as shown in \cref{sec:appendix-exp-other}) and also adaptable to noisy weak supervision $\hat{W}$ with an inherent learnable noise model $P(W|\hat{W};\theta)$ in the EM, which is left for future work.
NFA minimization and determination techniques can also be applied to further reduce the complexity of the proposed algorithm, by decreasing the number of states $|Q|$ of NFAs.

\section{Conclusion}
\label{sec:conclusion}

In this paper, we demonstrated a general framework for learning from arbitrary weak supervision that unifies various forms of weak supervision and can be extended to more settings flexibly, including instance partial labels, aggregate observations, pairwise observations, and unlabeled data, which addresses a significant gap in the practical applicability and scalability of weakly supervised learning methods.
Experiments across various settings and practical datasets validated the superiority of the proposed method.
We hope our work can inspire more research on weak supervision. 

\section*{Impact Statement}
This paper presents a unified framework for learning with arbitrary weak supervision. 
It has the potential to be broadly applied to many weakly-supervised learning settings in practice, advancing their deployment in industry and academia.

\section*{Acknowledge}
Masashi Sugiyama was supported by Institute for AI and Beyond, The University of Tokyo.



\bibliography{ref}
\bibliographystyle{icml2024}

\newpage
\appendix
\onecolumn

\section{Proofs}

\subsection{Derivation of \cref{eq:em}}
\label{sec:appendix-proof-theorm-em}

Evidence lower bound (ELBO), or equivalently variational lower bound \citep{dempster1977maximum}, is the core quantity in EM. 
We provide the detailed derivation for \cref{eq:em} here. 
To model $\log P(X, W;\theta)$:
\begin{equation}
\begin{split}
\log P(X, W; \theta) &= \log \sum_{Y} P(X, W, Y;\theta) \\
&= \log Q(Y) \frac{P(X, W, Y;\theta)}{Q(Y)} \\
&= \log \mathbb{E}_{Q(Y)} [ \frac{P(X, W, Y;\theta)}{Q(Y)} ] \\
&\geq \mathbb{E}_{Q(Y)} [\log  \frac{P(X, W, Y;\theta)}{Q(Y)} ] \quad \text{\small{Jensen’s inequality}} \\
&= \mathbb{E}_{Q(Y)} [\log P(X, W, Y;\theta)] - \mathbb{E}_{Q(Y)} [\log Q(Y)] \\
&= ELBO(\theta, Q(Y)), \\
\end{split}
\end{equation}
where the first term in ELBO is the lower bound and the second term is the entropy over $Q(Y)$ that is independent of $\theta$. 
Given the ELBO, we also have:
\begin{equation}
\begin{split}
    ELBO(\theta, Q(Y)) &= \mathbb{E}_{Q(Y)} [\log  \frac{P(X, W, Y;\theta)}{Q(Y)} ] \\
    &=  \mathbb{E}_{Q(Y)} [\log \frac{P(X, W, Y;\theta) P(Y|X, W;\theta)}{Q(Y) P(Y|X, W;\theta) } ] \\
    &=  \mathbb{E}_{Q(Y)} [\log \frac{P(Y| X, W;\theta) P(X, W; \theta) P(Y|X, W;\theta)}{Q(Y) P(Y|X, W;\theta) } ] \\
    &= \mathbb{E}_{Q(Y)} [\log P(X, W;\theta) \frac{P(Y|X, W;\theta)}{Q(Y)} ] \\
    &= \mathbb{E}_{Q(Y)} [\log P(X, W;\theta) ] - \mathbb{E}_{Q(Y)}[\frac{Q(Y)}{ P(Y|X, W;\theta)}] \\
    &= \log P(X, W;\theta) - \mathrm{KL}(Q(Y) || P(Y|X,W;\theta)).
\end{split}
\end{equation}
Thus we can see that maximizing the ELBO is equivalent to maximizing $\log P(X, W;\theta)$ when $P(Y|X, W;\theta)$ is close to $Q(Y)$, i.e., the Kullback-Leibier divergence $\mathrm{KL}(Q(Y) || P(Y|X,W;\theta))$ is approaching to 0. 
Thus we take $Q(Y) = P(Y|X, W;\theta^t)$ with current estimation $\theta^t$ from the model, and obtain \cref{eq:em}.

\subsection{Proof of \cref{proposition:em-loss}}
\label{sec:appendix-proof-proposition-em-loss}

\begin{proof}
Applying the maximum log-likelihood estimation to the weak supervision dataset $\mathcal{D} = \{(\mathbf{x}_i^{1:L}, w_i)\}_{i \in [N]}$, where $w \in W$ is the weak supervision for each sequence input $\mathbf{x}^{1:L}$ with $L \geq 1$. 
When $L=1$, $\mathbf{x}$ represents an individual training instance, otherwise it represents a group of sequence as discussed in the main paper. 
For simplicity, we consider $L$ as a fixed value for $\mathcal{L}$ here, but in practice it can denote variable length. 
We have \cref{assump} that the predictions and precise labels in the sequence are conditionally independent given whole input sequence. 

\begin{equation}
\begin{split}
    & \argmax_{\theta} \mathbb{E}_{Y|X,W;\theta^t} [\log P(X, W, Y;\theta)]  \\ 
    = & \argmax_{\theta} \mathbb{E}_{Y|X,W;\theta^t}  [\log P(W | Y, X;\theta) P(Y|X;\theta) P(X; \theta) ] \\
    = &\argmax_{\theta} \mathbb{E}_{Y|X,W;\theta^t} [\log P(Y|X;\theta)] +  \mathbb{E}_{Y|X,W;\theta^t}  [\log P(W | Y, X;\theta)] \quad \text{\small{ $P(X)$ is independent of $\theta$}} \\
    = & \argmax_{\theta}  \mathbb{E}_{Y|X,W;\theta^t} [\log P(Y|X;\theta)] + \log P(W | Y, X;\theta) \quad \text{\small{ $P(W | Y, X;\theta)$ is fixed for any $P(Y|X,W;\theta^t)$ }} 
\end{split}
\end{equation}
The derived objective $\mathcal{L}_{\mathrm{Weak}}$ on the dataset $\mathcal{D}$ from the EM formulation thus have two terms, where the first unsupervised term $\mathcal{L}_{\mathrm{U}}$ corresponds to:
\begin{equation}
\begin{split}
    \mathcal{L}_{\mathrm{U}} &= \sum_{i=1}^N \mathbb{E}_{y_i^{1:L} | \mathbf{x}_i^{1:L}, w_i;\theta^t} [\log p(y_i^{1:L} | \mathbf{x}_i^{1:L} ;\theta ) ] \\
    &=  \sum_{i=1}^N p(y_i^{1:L} | \mathbf{x}_i^{1:L}, w_i;\theta^t) \log \prod_{j=1}^L p(y_i^j | \mathbf{x}_i^j;\theta) \quad \text{\small{Instance-level model}} \\
    &=  \sum_{i=1}^N \sum_{j=1}^L p(y_i^j | \mathbf{x}_i^{1:L}, w_i;\theta^t) \log p(y_i^j | \mathbf{x}_i^j;\theta) \quad \text{\small{Conditional independence in \cref{assump}}},
\end{split}
\end{equation}
and the supervised term $\mathcal{L}_{\mathrm{S}}$ as:
\begin{equation}
    \mathcal{L}_{S} = \sum_{i=1}^N \log p(w_i | y_i^{1:L}, \mathbf{x}_i^{1:L};\theta)
\end{equation}
\end{proof}



\section{Method}
\label{sec:appendix-method}

\subsection{Illustration of Possible Labelings as Trellis of Common Weak Supervision Settings}
\label{sec:appendix-method-trellis}

Here, we present more illustration of the expanded trellis from the NFA in \cref{fig:nfa}. 
The demonstration of weak supervision is over a group of 4 instances for LProp and 2 instances for pairwise observations.

\begin{figure}[h!]
    \centering
    \includegraphics[width=0.5\textwidth]{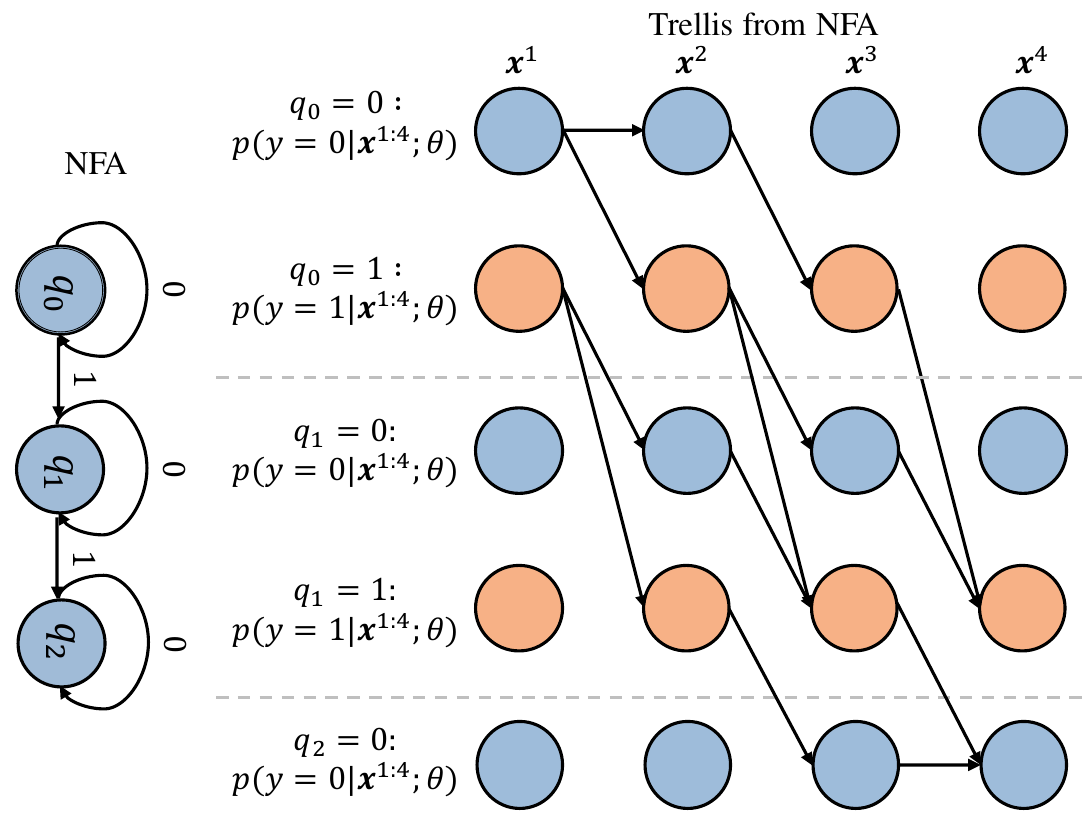}
    \caption{Illustration of trellis expanded from the NFA of label proportion on 4 instances whose weak supervision is exact two positive samples. We omit the last state of $q_2 = 1$ for simplicity because no path goes through it.}
    \label{fig:appendix-method-trellis-lprop}
\end{figure}

The trellis of LProp with exact two positive samples is shown in \cref{fig:appendix-method-trellis-lprop}. 
Note that for LProp, the number of states in its NFA depends on the exact count from the weak supervision as discussed in the main paper. 
The unlabeled data with class prior information can also be represented as expected label count and uses the trellis representation of LProp.

\begin{figure}[h!]
    \centering
    \includegraphics[width=0.4\textwidth]{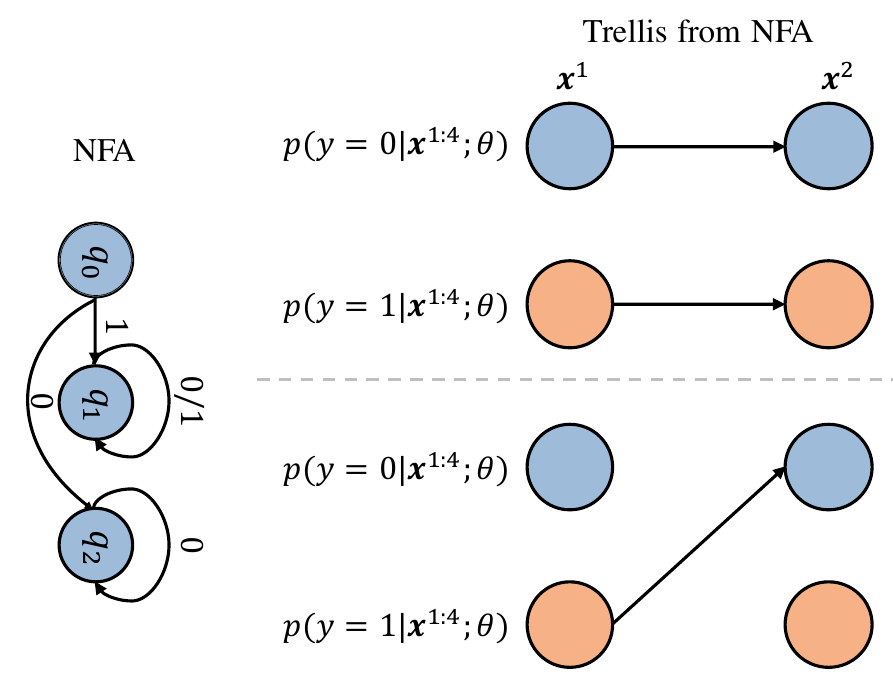}
    \caption{Illustration of trellis expanded from the NFA of pairwise comparison on 2 instances whose weak supervision is the first instance is more position than the second. We directly use 4 states to fully represent the cases $\{(0,0), (1, 1), (0, 1), (1, 0)\}$, which might looks different from its NFA who has only 3 states, but they indicate the same weak supervision.}
    \label{fig:appendix-method-trellis-pcomp}
\end{figure}

\begin{figure}[h!]
    \centering

    \hfill
    \subfigure[Pairwise similarity]{\label{fig:appendix-method-trellis-psim}\includegraphics[width=0.4\textwidth]{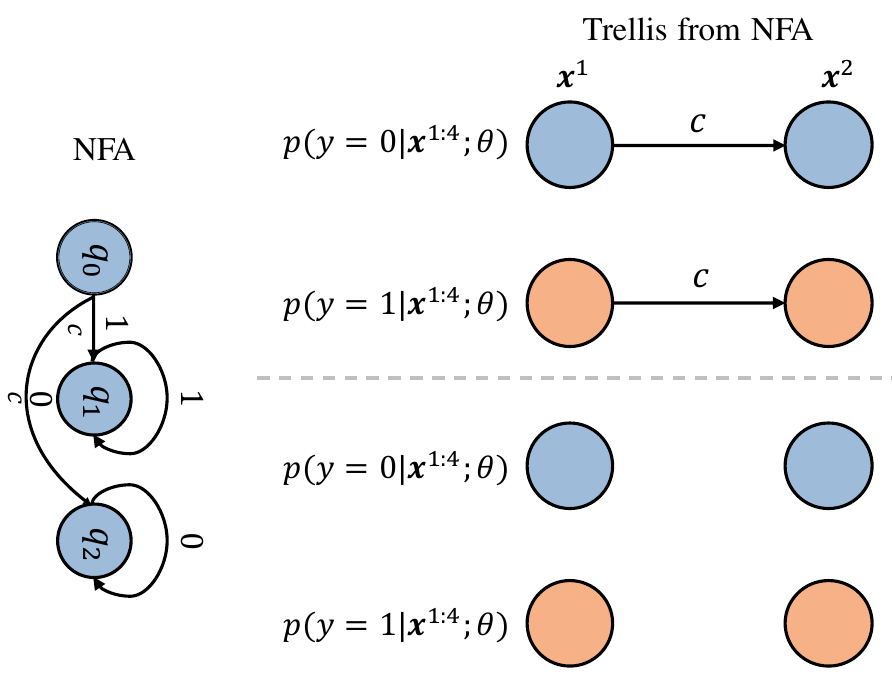}}
    \hfill
    \subfigure[Pairwise dissimilarity]{\label{fig:appendix-method-trellis-pdsim}\includegraphics[width=0.4\textwidth]{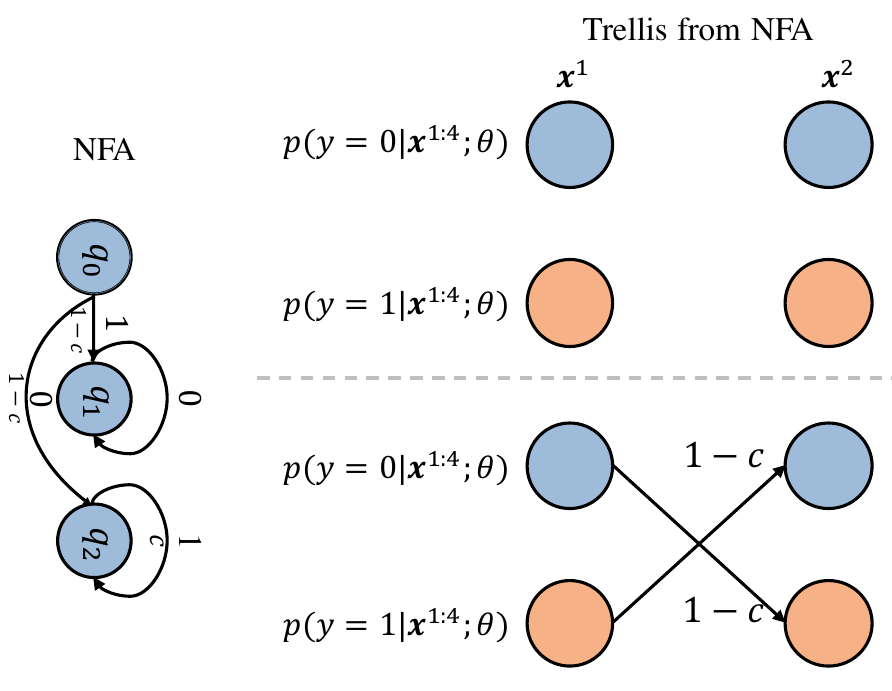}}
    \hfill
    
    \caption{Illustration of trellis expanded from the NFA of (a) pairwise similarity and (b) pairwise dissimilarity on 2 instances whose weak supervision is the whether the pair has similar or dissimilar supervision. We directly use 4 states to fully represent the cases $\{(0,0), (1, 1), (0, 1), (1, 0)\}$, which might looks different from its NFA who has only 3 states, but they indicate the same weak supervision. Similarity confidence and confidence different can also be represented using the trellis here by weighting each path according to the similarity or confidence score.}
    \label{fig:appendix-method-trellis-psim-pdsim}
\end{figure}

We also present the illustration of PComp, PSim, and PDsim in \cref{fig:appendix-method-trellis-pcomp}, \cref{fig:appendix-method-trellis-psim}, and \cref{fig:appendix-method-trellis-pdsim} respectively. 
Although we use 3 states in their NFA, we instead directly use 4 states in the expanded trellis to represent all the labelings for pairwise observations, i.e., $\{(0,0), (1, 1), (0, 1), (1, 0)\}$. 
Despite the notation difference, they represent the same weak supervision.
SimConf and ConfDiff can also be represented similarly by weighting the path with confidence score and similarity score.

For totally unlabeled data, every symbol in $\mathcal{Y}$ can be allowed for transition, thus its trellis degenerate to the prediction probability of each instance.

\subsection{Pseudo-algorithm of the Forward-Backward Algorithm of Common Weak Supervision Settings}
\label{sec:appendix-method-alg}

We present the pseudo-algorithm of performing the forward-backward algorithms on common weak supervision settings we evaluated. The pseudo-algorithm also corresponds to description of the trellis expanded from the NFA. 
Note that the only difference for each weak supervision setting is the NFA modeling. 
Once having the NFA modeling of weak supervision, the finite states and the transition between states are determined, and thus the forward-backward algorithm can be performed accordingly. 
We perform the forward-backward algorithm in log-space for numerical stability. 
Moreover, we use the log-sum-exp trick for computing the addition in log-space. 
For illustration simplicity, we present the pseudo-algorithm on single instance/group inputs and binary predictions, but in practice we implement the forward-backward pass at batch of instances/groups inputs and multi-class predictions. 
Here we illustrate the pseudo-algorithm for MultiIns in \cref{alg:multi_ins}, LProp in \cref{alg:label_prop}, PComp in \cref{alg:pcomp}, respectively. 
Other settings should either be similar or simple to solve.

\begin{algorithm}[h!]
\caption{Forward-Backward Algorithm for multiple instance (MultiIns) Learning}
\small
\begin{algorithmic}[1]
\REQUIRE Predicted probability in log-space as $log\_probs$ from $\mathbf{x}^{1:L}$, $w$ with 0 for no positive and 1 for at least one positive in the bag, bag length as $L$.
\STATE Number of states as $Q \gets 4$.
\STATE Initialize $\alpha \in \mathbb{R}^{2Q \times L}$ with $-1e12$ for forward pass.
\STATE $\alpha[0, 0], \alpha[1, 0] \gets log\_probs[0, 0], log\_probs[0, 1]$.
\FOR{$i = 1$ to $L$}
    \IF{$i < L - 1$}
        \STATE Update $\alpha[0, i] = \alpha[0, i - 1] + log\_probs[i, 0]$.
    \ELSE
        \STATE Update $\alpha[0, i] = -1e12$.
    \ENDIF
    \IF{$i \geq 2$}
        \STATE Update $\alpha[2, i] = \alpha[1, i - 1] + \alpha[2, i - 1] + \alpha[3, i - 1] + log\_probs[i, 0]$.
        \STATE Update $\alpha[3, i] = \alpha[1, i - 1] + \alpha[2, i - 1] + \alpha[3, i - 1] + log\_probs[i, 1]$.
    \ELSE
        \STATE Update $\alpha[2, i] = \alpha[1, i - 1] + log\_probs[i, 0]$.
        \STATE Update $\alpha[3, i] = \alpha[1, i - 1] + log\_probs[i, 1]$.
    \ENDIF
\ENDFOR
\STATE Compute forward probability $p(w|\mathbf{x}^{1:L},y^{1:L};\theta)$ from $\exp(\alpha)$ as $sup\_preds$.
\IF{w = 0}
    \STATE $em\_targets = ones\_like(log\_probs)$  
    \STATE Return $sup\_preds$, $em\_targets$ 
\ENDIF 
\STATE Initialize $\beta \in \mathbb{R}^{2Q \times L}$ with $-1e12$ for backward pass.
\STATE $\beta[1, L - 1], \beta[2, L - 1], \beta[3, L - 1] \gets log\_probs[L - 1, 1], log\_probs[L - 1, 0], log\_probs[L - 1, 1]$.
\FOR{$i = L-2$ down to $0$}
    \STATE $\beta[0, i] = \beta[0, i + 1] + \beta[1, i + 1] + log\_probs[i, 0]$.
    \STATE $\beta[1, i] = \beta[2, i + 1] + \beta[3, i + 1] + log\_probs[i, 1]$.
    \IF{$i > 0$}
        \STATE $\beta[2, i] = \beta[2, i + 1] + \beta[3, i + 1] + log\_probs[i, 0]$.
        \STATE $\beta[3, i] = \beta[2, i + 1] + \beta[3, i + 1] + log\_probs[i, 1]$.
    \ENDIF
\ENDFOR
\STATE Adjust $\beta$ based on $log\_probs$.
\STATE $\gamma = \alpha + \beta$.
\STATE $\gamma = \exp(\gamma.transpose(0, 1))$.
\STATE Compute joint probability $p(y^j | \mathbf{x}^{1:L}, w;\theta)$ as $em\_targets$
\STATE Return $sup\_preds$, $em\_targets$
\end{algorithmic}
\label{alg:multi_ins}
\end{algorithm}
\clearpage

\begin{algorithm}[h!]
\caption{Forward-Backward Algorithm for label proportion (LProp) Learning}
\small
\begin{algorithmic}[1]
\REQUIRE Predicted probability in log-space as $log\_probs$ of $\mathbf{x}^{1:L}$, $w$ indicates the count of positive instance.
\STATE Number of states $Q \gets 2 \times w + 1$.
\STATE Initialize $\alpha \in \mathbb{R}^{Q \times L}$ with $-1e12$ for forward pass.
\STATE $\alpha[0, 0] \gets log\_probs[0, 0]$.
\IF{$count > 0$}
    \STATE $\alpha[1, 0] \gets log\_probs[0, 1]$.
\ENDIF
\FOR{$i = 1$ to $L$}
    \STATE Update $\alpha[0, i] = \alpha[0, i - 1] + log\_probs[i, 0]$.
    \IF{$count > 0$}
        \STATE Update $\alpha[1, i] = \alpha[0, i - 1] + log\_probs[i, 1]$.
    \ENDIF
    \FOR{$j = 2$ to $Q$}
        \IF{$i < w - (Q - j) // 2$}
            \STATE Continue to next iteration of $j$.
        \ENDIF

        \IF{j \% 2 = 0}
            \STATE Update $\alpha[j, i] = \alpha[j, i - 1] + \alpha[j - 1, i - 1] + log\_probs[i, 0]$
        \ELSE
            \STATE Update $\alpha[j, i] = \alpha[j - 1, i - 1] + \alpha[j - 2, i - 1] + log\_probs[i, 1]$
        \ENDIF 
    \ENDFOR
\ENDFOR
\STATE Compute forward probability $sup\_preds$ from $\exp(\alpha)$.
\STATE Adjust $\alpha$ based on $w$ and $Q$ to avoid underflow.
\STATE Initialize $\beta \in \mathbb{R}^{Q \times L}$ with $-1e12$ for backward pass.
\STATE Set initial values of $\beta[-1, -1]$ and $\beta[-2, -1]$ based on $w$.
\FOR{$i = b - 2$ down to $0$}
    \IF{$i \geq w$}
        \STATE $\beta[-1, i] = \beta[-1, i + 1] + log\_probs[i, 0]$
    \ENDIF 
    
    \IF{$i \geq w  \& w > 0$}
        \STATE $\beta[-2, i] = \beta[-1, i + 1] + log\_probs[i, 0]$
    \ENDIF 
    
    \FOR{$j = 0$ to $k - 2$}
        \IF{$i < count - (k - j) // 2$}
            \STATE Continue to next iteration of $j$.
        \ENDIF

        \IF{j \% 2 = 0}
            \STATE Update $\beta[j, i] = \beta[j, i + 1] + \beta[j + 1, i + 1] + log\_probs[i, 0]$
        \ELSE
            \STATE Update $\beta[j, i] = \beta[j + 1, i + 1] + \beta[j + 2, i + 1] + log\_probs[i, 1]$
        \ENDIF 

    \ENDFOR
\ENDFOR
\STATE Adjust $\beta$ based on $log\_probs$.
\STATE $\beta = \exp(\beta)$
\STATE $\gamma = \alpha + \beta$.
\STATE $\gamma = \exp(\gamma.transpose(0, 1))$.
\STATE Compute joint probability $p(y^j | \mathbf{x}^{1:L}, w;\theta)$ as $em\_targets$
\STATE Return $sup\_preds$, $em\_targets$
\end{algorithmic}
\label{alg:label_prop}
\end{algorithm}

\clearpage

\begin{algorithm}
\caption{Forward-Backward Algorithm for pairwise comparison (PComp)  Learning}
\begin{algorithmic}[1]
\REQUIRE Predicted probability in log-space as $log\_probs$ of $\mathbf{x}^{1:L}$.
\STATE Number of states $Q \gets 4$.
\STATE Initialize $\log\_alpha \in \mathbb{R}^{4 \times 2}$ with $-1e12$ for the forward pass.
\STATE $\alpha[0, 0] = log\_probs[0, 0]$
\STATE $\alpha[1, 0] = log\_probs[0, 1]$
\STATE $\alpha[3, 0] = log\_probs[0, 1]$
\STATE $\alpha[0, 1] = \alpha[0, 0] + log\_probs[1, 0]$
\STATE $\alpha[1, 1] = \alpha[1, 0] + log\_probs[1, 1]$
\STATE $\alpha[2, 1] = \alpha[3, 0] + log\_probs[1, 0]$
\STATE Compute forward probability from $\exp(\alpha)$.
\STATE Initialize $\log\_beta \in \mathbb{R}^{4 \times 2}$ with $-1e12$ for the backward pass.
\STATE $\beta[0, 1] = log\_probs[1, 0]$
\STATE $\beta[1, 1] = log\_probs[1, 1]$
\STATE $\beta[2, 1] = log\_probs[1, 0]$
\STATE $\beta[0, 0] = \beta[0, 1] + log\_probs[0, 0]$
\STATE $\beta[1, 0] = \beta[1, 1] + log\_probs[0, 1]$
\STATE $\beta[3, 0] = \beta[1, 2] + log\_probs[0, 1]$
\STATE Adjust $\beta$ based on repeated $log\_probs$.
\STATE $\gamma = \alpha + \beta$.
\STATE $\gamma = \exp(\gamma.transpose(0, 1))$.
\STATE Compute the EM targets $em\_targets$ from $\gamma$.
\STATE Return $em\_targets$, $sup\_preds$
\end{algorithmic}
\label{alg:pcomp}
\end{algorithm}

\clearpage

\section{Experiments}
\label{sec:appendix-exp}

In this section, we provide more details on the training setup and hyper-parameters for our evaluations. We also present the details on datasets and class split of the datasets. 
More results of other weak supervision settings can be found in \cref{sec:appendix-exp-other}.

\subsection{Datasets and Classes Splits}
\label{sec:appendix-data}

\begin{table}[h!]
\caption{Dataset details}
\centering
\resizebox{0.5 \columnwidth}{!}{%
\begin{tabular}{@{}ccccc@{}}
\toprule
Dataset      & \# Classes & \# Training & \# Validation & \# Unlabeled \\ \midrule
MNIST        & 10         & 60,000      & 10,000        & -            \\
F-MNIST      & 10         & 60,000      & 10,000        & -            \\
CIFAR-10     & 10         & 50,000      & 10,000        & -            \\
CIFAR-100    & 100        & 50,000      & 10,000        & -            \\
STL-10       & 10         & 5,000       & 8,000         & 100,000      \\
ImageNet-100 & 100        & 130,000     &        5,000       & -            \\ \bottomrule
\end{tabular}%
}
\label{tab:appendix-dataset}
\end{table}

The datasets details are shown in \cref{tab:appendix-dataset}.

For some weak supervision settings, such as pairwise observations, positive unlabeled, and unlabeled unlabeled learning, we split the classes of each dataset into binary as follows.

\textbf{MNIST}. For multiple instance learning and label proportion learning, we set digit 9 as positive class, and others as negative class for binary classification. For other settings, we set digits 0-4 as positive class, and others as negative class. 

\textbf{F-MNIST}. Similarly, for multiple instance learning and label proportion learning, we set the 9-th class as positive class. For other settings, we set the classes related to tops as positive class, i.e., $\{5, 7, 9\}$.

\textbf{CIFAR-10 and STL-10}. For multiple instance learning and label proportion learning, we set bird, i.e., class 3, as positive class. For other settings, we set transportation related classes as positive class, i.e., airplane, automobile, ship, truck.

\textbf{CIFAR-100}. Binary classification on CIFAR-100 is not conduced on multiple instance learning and label proportion learning. For other settings, we select the 40 animal related classes from 100 total classes as positive class.

\subsection{Partial Labels}

Here we provide more training details and results of partial label learning.

\subsubsection{Setup}
\label{sec:appendix-exp-partial-setup}

We follow RCR \cite{revisitpllwu22l} for experiments of partial label learning. 
More specifically, we generate synthetic uniform partial label datasets, where we uniformly select each incorrect label for each instance into a candidate label set with partial ratio as probability. 
We adopt same training hyper-parameters for the baseline methods and \method for fair comparison.
A summarize of training parameters is shown in \cref{tab:append-param-pll}.

\begin{table}[h!]
\centering
\caption{Hyper-parameters for partial label (PartialL) learning used in experiments.}
\label{tab:append-param-pll}
\resizebox{0.7 \textwidth}{!}{%
\begin{tabular}{@{}c|ccccc@{}}
\toprule
\multicolumn{1}{c|}{Hyper-parameter}   & MNIST \& F-MNIST & CIFAR-10 & CIFAR-100 & STL-10 & ImageNet-100\\ \midrule
Image Size & 28 & 32 & 32 & 96 & 224 \\
Model      &  LeNet-5 & WRN-34-10 &   WRN-34-10         &    ResNet-18 &  ResNet-34   \\
Batch Size        &     64     &    64       &   64 &  64 & 32     \\
Optimizer & SGD & SGD & SGD & AdamW & AdamW \\
Learning Rate     &     0.1     &   0.1        &   0.1 & 0.001 & 0.001     \\
Weight Decay      &     1e-4     &   1e-4       &   1e-4    & 1e-4 & 1e-4  \\
LR Scheduler          &     MultiStep     &   MultiStep   &   MultiStep  & Cosine & Cosine     \\
Training Epochs   &     200    &   200        &  200 & 200 & 200       \\
\bottomrule
\end{tabular}%
}
\end{table}

For MNIST and F-MNIST, we use LeNet-5 \cite{lecun1998gradient}. 
We adopt WideResNet-34-10 variant \cite{zagoruyko2016wide} for CIFAR-10 and CIFAR-100, ResNet-18 \cite{he2016deep} for STL-10, and ResNet-34 for ImageNet-100. 
For optimizer, we use SGD \cite{loshchilov2016sgdr} for MNIST, F-MNIST, CIFAR-10, CIFAR-100, and AdamW \cite{kingma2014adam} for STL-10 and ImageNet-100.

\subsubsection{Results}
\label{sec:appendix-exp-partial-results}

We present more results on partial label learning in \cref{tab:appendix-partial}, where our method in general achieves the best performance. 

\begin{table*}[t!]
\centering
\caption{Accuracy on partial label (PartialL) learning. All results are averaged over three runs. This table is complementary to \cref{tab:partial-label}.}
\label{tab:appendix-partial}
\resizebox{\textwidth}{!}{%
\begin{tabular}{@{}l|cccc|cccc|cccc|cccc|cc|cc@{}}
\toprule
Dataset &
  \multicolumn{4}{c|}{MNIST} &
  \multicolumn{4}{c|}{F-MNIST} &
  \multicolumn{4}{c|}{CIFAR-10} &
  \multicolumn{4}{c|}{CIFAR-100} &
  \multicolumn{2}{c|}{STL-10} &
  \multicolumn{2}{c}{ImageNet-100} \\ \midrule
Partial Ratio &
  0.10 &
  0.30 &
  0.50 &
  0.70 &
  0.10 &
  0.300 &
  0.50 &
  0.70 &
  0.10 &
  0.30 &
  0.50 &
  0.70 &
  0.01 &
  0.05 &
  0.10 &
  0.20 &
  0.10 &
  0.30 &
  0.01 &
  0.05 \\ \midrule
CC &
   99.25\scriptsize{$\pm$0.02} &
   99.18\scriptsize{$\pm$0.05} &
   99.08\scriptsize{$\pm$0.03} &
   98.93\scriptsize{$\pm$0.05} &
   \textbf{91.44\scriptsize{$\pm$0.16}} &
   91.10\scriptsize{$\pm$0.07} &
   90.45\scriptsize{$\pm$0.09} &
   89.55\scriptsize{$\pm$0.28} &
   95.25\scriptsize{$\pm$0.03} &
   94.13\scriptsize{$\pm$0.09} &
   92.51\scriptsize{$\pm$0.04} &
   89.01\scriptsize{$\pm$0.20} &
   79.68\scriptsize{$\pm$0.14} &
   78.73\scriptsize{$\pm$0.24} &
   77.44\scriptsize{$\pm$0.32} &
   74.60\scriptsize{$\pm$0.17} &
   77.02\scriptsize{$\pm$0.69} &
   73.26\scriptsize{$\pm$0.34} &
   73.14\scriptsize{$\pm$0.94} &
   64.67\scriptsize{$\pm$0.74} \\
LWS &
   98.23\scriptsize{$\pm$0.02} &
   98.04\scriptsize{$\pm$0.12} &
   97.95\scriptsize{$\pm$0.11} &
   96.96\scriptsize{$\pm$0.10} &
   88.17\scriptsize{$\pm$0.11} &
   88.10\scriptsize{$\pm$0.05} &
   87.59\scriptsize{$\pm$0.18} &
   86.60\scriptsize{$\pm$0.13} &
   91.42\scriptsize{$\pm$0.03} &
   88.76\scriptsize{$\pm$0.45} &
   85.66\scriptsize{$\pm$0.32} &
   80.71\scriptsize{$\pm$0.10} &
   69.46\scriptsize{$\pm$0.28} &
   55.49\scriptsize{$\pm$0.67} &
   50.67\scriptsize{$\pm$0.33} &
   43.51\scriptsize{$\pm$0.32} &
   67.65\scriptsize{$\pm$0.33} &
   58.18\scriptsize{$\pm$1.65} &
   72.04\scriptsize{$\pm$0.77} &
   62.13\scriptsize{$\pm$0.95} \\
PRODEN &
   99.12\scriptsize{$\pm$0.52} &
   98.89\scriptsize{$\pm$0.52} &
   98.27\scriptsize{$\pm$0.68} &
   97.77\scriptsize{$\pm$0.82} &
   90.95\scriptsize{$\pm$0.63} &
   \textbf{91.96\scriptsize{$\pm$0.70}} &
   90.40\scriptsize{$\pm$0.58} &
   89.20\scriptsize{$\pm$0.45} &
   95.25\scriptsize{$\pm$0.45} &
   95.68\scriptsize{$\pm$0.40} &
   93.85\scriptsize{$\pm$0.60} &
   91.11\scriptsize{$\pm$0.70} &
   79.06\scriptsize{$\pm$0.24} &
   79.17\scriptsize{$\pm$0.36} &
   77.80\scriptsize{$\pm$0.31} &
   74.99\scriptsize{$\pm$0.57} &
   77.74\scriptsize{$\pm$0.52} &
   74.18\scriptsize{$\pm$0.41} &
   78.61\scriptsize{$\pm$0.63} &
   77.59\scriptsize{$\pm$0.60} \\
PiCO &
   99.22\scriptsize{$\pm$0.01} &
   99.20\scriptsize{$\pm$0.01} &
   99.10\scriptsize{$\pm$0.02} &
   98.96\scriptsize{$\pm$0.09} &
   90.30\scriptsize{$\pm$1.44} &
   91.41\scriptsize{$\pm$0.05} &
   90.42\scriptsize{$\pm$0.14} &
   89.73\scriptsize{$\pm$0.21} &
   95.37\scriptsize{$\pm$0.12} &
   95.14\scriptsize{$\pm$0.16} &
   93.32\scriptsize{$\pm$0.23} &
   90.26\scriptsize{$\pm$0.20} &
   79.49\scriptsize{$\pm$0.13} &
   78.71\scriptsize{$\pm$0.18} &
   77.50\scriptsize{$\pm$0.15} &
   74.89\scriptsize{$\pm$0.13} &
   77.44\scriptsize{$\pm$0.26} &
   73.19\scriptsize{$\pm$1.05} &
   80.93\scriptsize{$\pm$0.81} &
   78.74\scriptsize{$\pm$1.34} \\
RCR &
   99.25\scriptsize{$\pm$0.04} &
   99.21\scriptsize{$\pm$0.04} &
   99.11\scriptsize{$\pm$0.03} &
   99.01\scriptsize{$\pm$0.05} &
   91.26\scriptsize{$\pm$0.17} &
   91.26\scriptsize{$\pm$0.08} &
   90.82\scriptsize{$\pm$0.12} &
   90.06\scriptsize{$\pm$0.03} &
   95.57\scriptsize{$\pm$0.19} &
   94.65\scriptsize{$\pm$0.05} &
   94.04\scriptsize{$\pm$0.02} &
   91.45\scriptsize{$\pm$0.10} &
   79.89\scriptsize{$\pm$0.23} &
   78.93\scriptsize{$\pm$0.30} &
   78.03\scriptsize{$\pm$0.07} &
   75.40\scriptsize{$\pm$0.12} &
   78.02\scriptsize{$\pm$0.40} &
   74.67\scriptsize{$\pm$0.56} &
   81.52\scriptsize{$\pm$0.94} &
   79.67\scriptsize{$\pm$1.22} \\
\cellc  \method &
 \cellc  \textbf{99.25\scriptsize{$\pm$0.01}} &
 \cellc   \textbf{99.28\scriptsize{$\pm$0.05}} &
 \cellc   \textbf{99.12\scriptsize{$\pm$0.02}} &
 \cellc   \textbf{99.14\scriptsize{$\pm$0.04}} &
 \cellc   91.42\scriptsize{$\pm$0.22} &
  \cellc  91.28\scriptsize{$\pm$0.09} &
  \cellc  \textbf{90.85\scriptsize{$\pm$0.10}} &
  \cellc  \textbf{90.35\scriptsize{$\pm$0.15}} &
   \cellc \textbf{95.61\scriptsize{$\pm$0.03}} &
   \cellc \textbf{95.23\scriptsize{$\pm$0.11}} &
  \cellc  \textbf{94.31\scriptsize{$\pm$0.09}} &
  \cellc  \textbf{92.06\scriptsize{$\pm$0.14}} &
 \cellc   \textbf{80.06\scriptsize{$\pm$0.17}} &
 \cellc   \textbf{79.47\scriptsize{$\pm$0.09}} &
 \cellc   \textbf{78.35\scriptsize{$\pm$0.11}} &
  \cellc  \textbf{75.82\scriptsize{$\pm$0.25}} &
 \cellc   \textbf{78.56\scriptsize{$\pm$0.27}} &
 \cellc   \textbf{74.79\scriptsize{$\pm$0.21}} &
 \cellc   \textbf{82.66\scriptsize{$\pm$0.54}} &
  \cellc  \textbf{81.09\scriptsize{$\pm$0.50}} \\
\bottomrule
\end{tabular}%
}
\end{table*}

\subsection{Aggregate Observations}

More details about experiments of aggregate observations are shown here. 

\subsubsection{Setup}
\label{sec:appendix-exp-aggre-setup}

For aggregate observation, the largest dataset previously experimented is MNIST, which is unpractical. Here we present the training hyper-parameters we used for MultiIns and LProp in \cref{tab:append-param-aggregate}.

\begin{table}[h!]
\centering
\caption{Hyper-parameters for multiple instance (MultiIns) and label proportion (LProp) learning used in experiments.}
\label{tab:append-param-aggregate}
\resizebox{0.7 \textwidth}{!}{%
\begin{tabular}{@{}c|ccccc@{}}
\toprule
\multicolumn{1}{c|}{Hyper-parameter}   & MNIST \& F-MNIST & CIFAR-10 & CIFAR-100 & STL-10 & ImageNet-100\\ \midrule
Image Size & 28 & 32 & 32 & 96 & 224 \\
Model      &  LeNet-5 & WRN-28-2 &   ResNet-18        &    ResNet-18 &  ResNet-34   \\
Batch Size        &     4     &    4       &   4 &  4 & 8     \\
Optimizer & AdamW & AdamW & AdamW & AdamW & AdamW \\
Learning Rate     &     5e-4     &   1e-3        &  1e-3 &  1e-3 & 1e-3     \\
Weight Decay      &     1e-4     &   5e-4       &   5e-4    & 5e-4 & 1e-4  \\
LR Scheduler          &     Cosine     &   Cosine   &   Cosine  & Cosine & Cosine     \\
Training Epochs   &     100    &   100        &  100 & 100 & 100       \\
\bottomrule
\end{tabular}%
}
\end{table}

We train all methods in both settings for 100 epochs and AdamW optimizer. We set the learning rate to 1e-4 for MNIST and F-MNIST, and 1e-3 for others. 
WideResNet-28-2 is utlized for CIFAR-10, while ResNet-18 is used for CIFAR-100 and STL-10. 
Since each training instance for aggregate observations is a group of examples of variable length, we set batch size to $4$ universally or $8$ for ImageNet-100.

To create aggregate observations, we sample instances from the dataset to form groups/bags according to the specified Gaussian distribution. Then we summarize the weak supervision as counts of the labels in the group. which eventually convert to flags of existence of positive samples for multiple instance learning. 
For binary classification, we ensure that the number of negative bags and positive bags are balanced.

\subsubsection{Results}
\label{sec:appendix-exp-aggre-results}

\begin{table*}[t!]
\centering
\caption{Accuracy on both binary and multi-class multi-label aggregate observations of multiple instance (MI) learning and label proportion (LP) learning. All results are averaged over three runs. This table is complementary to \cref{tab:aggregate}.}
\label{tab:appendix-aggregate}
\resizebox{0.95 \textwidth}{!}{%
\begin{tabular}{@{}c|cccc|cccc|cc|cc@{}}
\toprule
  Dataset &
  \multicolumn{4}{c|}{MNIST} &
  \multicolumn{4}{c|}{F-MNIST} &
  \multicolumn{2}{c|}{CIFAR-10} &
  \multicolumn{2}{c}{STL-10} \\ \cmidrule(l){1-13} 
  \multicolumn{1}{l|}{\# Classes} &
  \multicolumn{2}{c}{2} &
  \multicolumn{2}{c|}{10} &
  \multicolumn{2}{c}{2} &
  \multicolumn{2}{c|}{10} &
  \multicolumn{2}{c|}{2} &
  \multicolumn{2}{c}{2} \\ \cmidrule(l){1-13} 
  Dist &
  $\mathcal{N}(10,2)$ &
  $\mathcal{N}(50, 10)$ &
  $\mathcal{N}(10,2)$ &
  $\mathcal{N}(20, 5)$ &
  $\mathcal{N}(10,2)$ &
  $\mathcal{N}(50, 10)$ &
  $\mathcal{N}(10,2)$ &
  $\mathcal{N}(20, 5)$ &
  $\mathcal{N}(10,2)$ &
  $\mathcal{N}(20, 5)$ &
  $\mathcal{N}(5, 1)$ &
  $\mathcal{N}(10,2)$ \\
  \# Bags &
  1,000 &
  250 &
  1,000 &
  500 &
  1,000 &
  250 &
  1,000 &
  500 &
  5,000 &
  2,500 &
  2,000 &
  1,000 \\ \midrule
  \midrule
  \multicolumn{13}{c}{Multiple Instance Learning} \\
  \midrule
  Count Loss &
  \textbf{97.05\scriptsize{$\pm$0.45}} &
  91.21\scriptsize{$\pm$0.46} &
  \textbf{97.61\scriptsize{$\pm$0.20}} &
  \textbf{94.90\scriptsize{$\pm$0.24}} &
  \textbf{97.64\scriptsize{$\pm$0.10}} &
  91.62\scriptsize{$\pm$1.96} &
  \textbf{86.24\scriptsize{$\pm$0.33}} &
  82.02\scriptsize{$\pm$0.06} &
  \textbf{63.07\scriptsize{$\pm$1.63}} &
  56.71\scriptsize{$\pm$2.23} &
  57.90\scriptsize{$\pm$6.11} &
  51.50\scriptsize{$\pm$1.65} \\
  UUM &
  81.08\scriptsize{$\pm$0.11} &
  74.00\scriptsize{$\pm$0.53} &
  63.96\scriptsize{$\pm$5.39} &
  23.43\scriptsize{$\pm$4.01} &
  91.40\scriptsize{$\pm$1.00} &
  87.38\scriptsize{$\pm$1.32} &
  64.24\scriptsize{$\pm$2.28} &
  28.57\scriptsize{$\pm$5.90} &
  58.25\scriptsize{$\pm$1.59} &
  57.67\scriptsize{$\pm$0.61} &
  57.05\scriptsize{$\pm$4.94} &
  57.60\scriptsize{$\pm$0.64} \\
  \cellc \method &
  \cellc 97.04\scriptsize{$\pm$0.38} &
  \cellc \textbf{91.54\scriptsize{$\pm$0.54}} &
  \cellc 97.57\scriptsize{$\pm$0.06} &
  \cellc 94.80\scriptsize{$\pm$0.27} &
  \cellc 97.59\scriptsize{$\pm$0.16} &
  \cellc \textbf{93.21\scriptsize{$\pm$1.74}} &
  \cellc 86.22\scriptsize{$\pm$0.18} &
  \cellc \textbf{82.05\scriptsize{$\pm$0.20}} &
  \cellc 62.63\scriptsize{$\pm$1.73} &
  \cellc \textbf{57.94\scriptsize{$\pm$2.11}} &
  \cellc \textbf{58.40\scriptsize{$\pm$3.11}} &
  \cellc \textbf{58.03\scriptsize{$\pm$0.73}} \\ \midrule

  \midrule
  \multicolumn{13}{c}{Label Proportion Learning} \\
  \midrule

  LLP-VAT &
  98.18\scriptsize{$\pm$0.19} &
  92.37\scriptsize{$\pm$2.15} &
  98.21\scriptsize{$\pm$0.07} &
  \textbf{98.41\scriptsize{$\pm$0.10}} &
  98.13\scriptsize{$\pm$0.10} &
  96.83\scriptsize{$\pm$0.11} &
  86.99\scriptsize{$\pm$0.45} &
  83.65\scriptsize{$\pm$0.94} &
  85.33\scriptsize{$\pm$0.44} &
  54.20\scriptsize{$\pm$2.72} &
  50.51\scriptsize{$\pm$0.36} &
  50.15\scriptsize{$\pm$0.21} \\
  Count Loss &
  \textbf{98.89\scriptsize{$\pm$0.21}} &
  96.46\scriptsize{$\pm$0.19} &
  97.95\scriptsize{$\pm$0.02} &
  98.29\scriptsize{$\pm$0.11} &
  \textbf{98.27\scriptsize{$\pm$0.12}} &
  \textbf{97.44\scriptsize{$\pm$0.16}} &
  87.50\scriptsize{$\pm$0.05} &
  85.70\scriptsize{$\pm$0.54} &
  89.46\scriptsize{$\pm$0.24} &
  67.58\scriptsize{$\pm$2.03} &
  65.93\scriptsize{$\pm$0.91} &
  56.23\scriptsize{$\pm$1.15} \\
  UUM &
  - &
  - &
  - &
  - &
  - &
  - &
  - &
  - &
  - &
  - &
  61.54\scriptsize{$\pm$2.16} &
  - \\
  \cellc \method &
  \cellc 98.62\scriptsize{$\pm$0.18} &
  \cellc \textbf{97.05\scriptsize{$\pm$0.13}} &
  \cellc \textbf{98.42\scriptsize{$\pm$0.11}} &
  \cellc 98.39\scriptsize{$\pm$0.08} &
  \cellc 98.18\scriptsize{$\pm$0.02} &
  \cellc 97.40\scriptsize{$\pm$0.12} &
  \cellc \textbf{88.02\scriptsize{$\pm$0.23}} &
  \cellc \textbf{86.20\scriptsize{$\pm$0.66}} &
  \cellc \textbf{89.77\scriptsize{$\pm$0.45}} &
  \cellc \textbf{68.03\scriptsize{$\pm$2.41}} &
  \cellc \textbf{66.04\scriptsize{$\pm$0.64}} &
  \cellc \textbf{58.20\scriptsize{$\pm$1.03}} \\ \bottomrule
\end{tabular}%
}
\end{table*}

We present more results of the binary classification of aggregate observations on MNIST, F-MNIST, CIFAR-10 and STL-10 in \cref{tab:appendix-aggregate}. 
The multi-class classification results of MNIST and F-MNIST are also shown here.
One can observe that, for both settings, our method is on par with Count Loss on MNIST and F-MNIST, and in general performs the best on multi-class classification settings of these two datasets.
Moreover, on binary classification of CIFAR-10 and STL-10, our method also outperforms the baselines.

\subsection{Pairwise Observations}

We provide more training details and results of pairwise observations here. 

\subsubsection{Setup}
\label{sec:appendix-exp-pair-setup}

\begin{table}[h!]
\centering
\caption{Hyper-parameters for pairwise comparison (PComp), pairwise similarity (PSim), similarity confidence (SimConf), and confidence difference (ConfDiff) learning used in experiments.}
\label{tab:append-param-pair}
\resizebox{0.55 \textwidth}{!}{%
\begin{tabular}{@{}c|cccc@{}}
\toprule
\multicolumn{1}{c|}{Hyper-parameter}   & MNIST \& F-MNIST & CIFAR-10 & CIFAR-100 & STL-10 \\ \midrule
Image Size & 28 & 32 & 32 & 96  \\
Model      &  LeNet-5 & WRN-28-2 &   ResNet-18        &    ResNet-18   \\
Batch Size        &     64     &    64       &   64 &  32     \\
Optimizer & AdamW & AdamW & AdamW & AdamW \\
Learning Rate     &     5e-4     &   1e-3        &  1e-3 &  1e-3     \\
Weight Decay      &     1e-4     &   1e-3       &   1e-3    & 1e-3  \\
LR Scheduler          &     Cosine     &   Cosine   &   Cosine  & Cosine     \\
Training Epochs   &     100    &   100        &  100 & 100      \\
\bottomrule
\end{tabular}%
}
\end{table}

For pairwise observations $(\mathbf{x}^1), \mathbf{x}^2)$, we adopt the same training parameters for the four settings we evaluated, as shown in \cref{tab:append-param-pair}.

For PComp, PSim, and SimConf of class prior $p$, we form the pair observations by sampling from all positive pairs following $p^2$, all negative pairs following $(1 - p)^2$, and positive and negative pairs following $2p(1-p)$, as in \citet{feng2021pointwise,uumwei23a,cao2021learning}.
For ConfDiff, we sample each instance in the pair independently according to the class prior $p$, as in \citet{wang2023binary}.
For PComp, the weak supervision is that $\mathbf{x}^1$ is more positive than $\mathbf{x}^2$. 
For PSim, the weak supervision is that the pairs are either similar or dissimilar. 
For SimConf and ConfDiff, we need pre-trained models to compute the similarity score as in \cite{cao2021learning} and \citet{wang2023binary} respectively. 
We set two pre-trained models. 
The first one is the same architecture shown in \cref{tab:append-param-pair}, trained on a separate set of instances in each dataset and used to compute the score for the sampled pairs.
The second one is CLIP models \cite{radford2021learning}, where we compute the scores in a zero-shot manner.

\subsubsection{Results}
\label{sec:appendix-exp-pair-results}

We present more results of PComp in \cref{tab:appendix-pcomp}, PSim in \cref{tab:appendix-psim}, SimConf in \cref{tab:appendix-simconf}, and ConfDiff in \cref{tab:appendix-confdiff}. 
Our method consistently and universally achieves the best performance on these settings in general.

\begin{table*}[t!]
\centering
\caption{Accuracy on pairwise comparison (PComp) learning for binary classification. All results are averaged over three runs.}
\label{tab:appendix-pcomp}
\resizebox{\textwidth}{!}{%
\begin{tabular}{@{}l|ccc|ccc|ccc|ccc|ccc@{}}
\toprule
Dataset &
  \multicolumn{3}{c|}{F-MNIST} &
  \multicolumn{3}{c|}{MNIST} &
  \multicolumn{3}{c|}{CIFAR-10} &
  \multicolumn{3}{c|}{CIFAR-100} &
  \multicolumn{3}{c}{STL-10} \\ \midrule
\#Pairs &
  \multicolumn{3}{c|}{25,000} &
  \multicolumn{3}{c|}{25,000} &
  \multicolumn{3}{c|}{20,000} &
  \multicolumn{3}{c|}{20.000} &
  \multicolumn{3}{c}{5,000} \\ \midrule
Prior &
  0.2 &
  0.5 &
  0.8 &
  0.2 &
  0.5 &
  0.8 &
  0.2 &
  0.5 &
  0.8 &
  0.2 &
  0.5 &
  0.8 &
  0.2 &
  0.5 &
  0.8 \\ \midrule
PComp ABS &
  92.82\scriptsize{$\pm$0.89} &
  99.73\scriptsize{$\pm$0.04} &
  90.96\scriptsize{$\pm$0.74} &
  91.54\scriptsize{$\pm$0.86} &
  96.86\scriptsize{$\pm$0.30} &
  91.09\scriptsize{$\pm$1.08} &
  88.75\scriptsize{$\pm$0.60} &
  91.78\scriptsize{$\pm$0.10} &
  87.37\scriptsize{$\pm$1.89} &
  73.10\scriptsize{$\pm$0.15} &
  81.67\scriptsize{$\pm$0.24} &
  66.06\scriptsize{$\pm$1.19} &
  78.38\scriptsize{$\pm$0.50} &
  79.07\scriptsize{$\pm$0.40} &
  56.45\scriptsize{$\pm$1.86} \\
PComp ReLU &
  99.65\scriptsize{$\pm$0.07} &
  99.73\scriptsize{$\pm$0.08} &
  98.41\scriptsize{$\pm$0.41} &
  90.30\scriptsize{$\pm$0.28} &
  96.71\scriptsize{$\pm$0.10} &
  92.87\scriptsize{$\pm$0.22} &
  90.47\scriptsize{$\pm$0.94} &
  92.18\scriptsize{$\pm$0.22} &
  90.57\scriptsize{$\pm$0.21} &
  73.10\scriptsize{$\pm$0.77} &
  81.77\scriptsize{$\pm$0.59} &
  66.57\scriptsize{$\pm$1.27} &
  \textbf{79.30\scriptsize{$\pm$0.85}} &
  79.68\scriptsize{$\pm$0.75} &
  67.01\scriptsize{$\pm$1.71} \\
PComp Teacher &
  92.41\scriptsize{$\pm$0.38} & 
  93.92\scriptsize{$\pm$0.81} &
  92.54\scriptsize{$\pm$0.15} &
  92.79\scriptsize{$\pm$0.45} &
  93.03\scriptsize{$\pm$0.93} &
  91.46\scriptsize{$\pm$1.31} &
  92.29\scriptsize{$\pm$0.19} &
  93.33\scriptsize{$\pm$0.38} &
  91.35\scriptsize{$\pm$0.27} &
  72.72\scriptsize{$\pm$0.33} &
  78.59\scriptsize{$\pm$0.60} &
  67.43\scriptsize{$\pm$3.09} &
  78.09\scriptsize{$\pm$0.68} &
  77.33\scriptsize{$\pm$0.14} &
  72.88\scriptsize{$\pm$0.15} \\
PComp Unbiased &
  87.64\scriptsize{$\pm$0.28} &
  89.30\scriptsize{$\pm$0.35} &
  81.16\scriptsize{$\pm$1.20} &
  76.23\scriptsize{$\pm$1.56} &
  84.35\scriptsize{$\pm$0.74} &
  78.81\scriptsize{$\pm$2.32} &
  88.13\scriptsize{$\pm$0.29} &
  91.71\scriptsize{$\pm$0.48} &
  88.22\scriptsize{$\pm$0.58} &
  66.02\scriptsize{$\pm$0.97} &
  67.80\scriptsize{$\pm$0.07} &
  60.86\scriptsize{$\pm$2.19} &
  76.85\scriptsize{$\pm$0.57} &
  77.46\scriptsize{$\pm$0.19} &
  71.60\scriptsize{$\pm$0.95} \\
Rank Pruning &
  90.32\scriptsize{$\pm$1.10} &
  91.93\scriptsize{$\pm$0.41} &
  89.99\scriptsize{$\pm$0.98} &
  90.56\scriptsize{$\pm$0.27} & 
  91.59\scriptsize{$\pm$1.31} &
  90.44\scriptsize{$\pm$0.69} &
  92.98\scriptsize{$\pm$0.30} &
  93.98\scriptsize{$\pm$0.40} &
  91.97\scriptsize{$\pm$0.27} &
  73.81\scriptsize{$\pm$1.21} &
  78.90\scriptsize{$\pm$0.48} &
  71.51\scriptsize{$\pm$0.73} &
  78.39\scriptsize{$\pm$0.33} &
  77.89\scriptsize{$\pm$0.42} &
  73.62\scriptsize{$\pm$1.38} \\
\cellc \method &
\cellc  \textbf{99.59\scriptsize{$\pm$0.01}} &
\cellc  \textbf{99.85\scriptsize{$\pm$0.02}} &
  \cellc \textbf{99.82\scriptsize{$\pm$0.03}} &
  \cellc \textbf{95.95\scriptsize{$\pm$0.15}} &
  \cellc \textbf{97.70\scriptsize{$\pm$0.11}} &
  \cellc \textbf{96.03\scriptsize{$\pm$0.39}} &
  \cellc \textbf{93.46\scriptsize{$\pm$0.32}} &
  \cellc \textbf{94.15\scriptsize{$\pm$0.10}} &
  \cellc \textbf{93.28\scriptsize{$\pm$0.38}} &
  \cellc \textbf{80.33\scriptsize{$\pm$0.07}} &
  \cellc \textbf{83.15\scriptsize{$\pm$0.16}} &
  \cellc \textbf{80.50\scriptsize{$\pm$0.20}} &
  \cellc 79.15\scriptsize{$\pm$0.78} &
  \cellc \textbf{81.26\scriptsize{$\pm$0.54}} &
  \cellc \textbf{79.24\scriptsize{$\pm$0.87}} \\ \bottomrule
\end{tabular}%
}
\end{table*}

\begin{table*}[t!]
\centering
\caption{Accuracy on pairwise similarity (PSim) learning for binary classification. All results are averaged over three runs.}
\label{tab:appendix-psim}
\resizebox{\textwidth}{!}{%
\begin{tabular}{@{}l|ccc|ccc|ccc|ccc|ccc@{}}
\toprule
Dataset &
  \multicolumn{3}{c|}{F-MNIST} &
  \multicolumn{3}{c|}{MNIST} &
  \multicolumn{3}{c|}{CIFAR-10} &
  \multicolumn{3}{c|}{CIFAR-100} &
  \multicolumn{3}{c}{STL-10} \\ \midrule
\#Pairs &
  \multicolumn{3}{c|}{30,000} &
  \multicolumn{3}{c|}{30,000} &
  \multicolumn{3}{c|}{25,000} &
  \multicolumn{3}{c|}{25.000} &
  \multicolumn{3}{c}{5,000} \\ \midrule
Prior &
  0.2 &
  0.4 &
  0.6 &
  0.2 &
  0.4 &
  0.6 &
  0.2 &
  0.4 &
  0.6 &
  0.2 &
  0.4 &
  0.6 &
  0.2 &
  0.4 &
  0.6 \\ \midrule
RiskSD &
  99.34\scriptsize{$\pm$0.17} &
  98.11\scriptsize{$\pm$0.11} &
  98.44\scriptsize{$\pm$0.45} &
  94.00\scriptsize{$\pm$0.61} &
  89.31\scriptsize{$\pm$0.65} &
  89.41\scriptsize{$\pm$0.58} &
  89.68\scriptsize{$\pm$0.67} &
  85.78\scriptsize{$\pm$1.70} &
  85.61\scriptsize{$\pm$1.34} &
  69.56\scriptsize{$\pm$3.00} &
  70.41\scriptsize{$\pm$0.21} &
  64.26\scriptsize{$\pm$3.81} &
  77.42\scriptsize{$\pm$0.94} &
  74.15\scriptsize{$\pm$3.27} &
  69.35\scriptsize{$\pm$0.32} \\
UUM &
  99.94\scriptsize{$\pm$0.01} &
  \textbf{99.93\scriptsize{$\pm$0.01}} &
  \textbf{99.93\scriptsize{$\pm$0.01}} &
  \textbf{99.04\scriptsize{$\pm$0.08}} &
  \textbf{99.12\scriptsize{$\pm$0.05}} &
  99.04\scriptsize{$\pm$0.13} &
  96.96\scriptsize{$\pm$0.20} &
  97.24\scriptsize{$\pm$0.23} &
  97.16\scriptsize{$\pm$0.24} &
  \textbf{86.95\scriptsize{$\pm$0.08}} &
  87.13\scriptsize{$\pm$0.40} &
  85.19\scriptsize{$\pm$2.45} &
  85.23\scriptsize{$\pm$1.06} &
  83.55\scriptsize{$\pm$0.80} &
  83.64\scriptsize{$\pm$0.25} \\
\cellc \method &
\cellc  \textbf{99.94\scriptsize{$\pm$0.01}} &
\cellc  \textbf{99.93\scriptsize{$\pm$0.01}} &
\cellc  \textbf{99.93\scriptsize{$\pm$0.01}} &
\cellc  98.96\scriptsize{$\pm$0.01} &
\cellc  99.07\scriptsize{$\pm$0.01} &
\cellc  \textbf{99.05\scriptsize{$\pm$0.10}} &
\cellc  \textbf{97.09\scriptsize{$\pm$0.04}} &
\cellc  \textbf{97.44\scriptsize{$\pm$0.07}} &
\cellc  \textbf{97.18\scriptsize{$\pm$0.22}} &
\cellc  86.89\scriptsize{$\pm$0.42} &
\cellc  \textbf{87.25\scriptsize{$\pm$0.16}} &
\cellc  \textbf{86.96\scriptsize{$\pm$0.33}} &
\cellc  \textbf{86.36\scriptsize{$\pm$1.60}} &
\cellc  \textbf{84.81\scriptsize{$\pm$0.60}} &
\cellc  \textbf{85.19\scriptsize{$\pm$0.26}} \\ \bottomrule
\end{tabular}%
}
\end{table*}

\begin{table*}[t!]
\centering
\caption{Accuracy on similarity confidence (SimConf) learning for binary classification. All results are averaged over three runs.}
\label{tab:appendix-simconf}
\resizebox{\textwidth}{!}{%
\begin{tabular}{@{}l|cccc|cccc|cccc|cccc|cccc@{}}
\toprule
Dataset &
  \multicolumn{4}{c|}{F-MNIST} &
  \multicolumn{4}{c|}{MNIST} &
  \multicolumn{4}{c|}{CIFAR-10} &
  \multicolumn{4}{c|}{CIFAR-100} &
  \multicolumn{4}{c}{STL-10} \\ \midrule
\#Pairs &
  \multicolumn{4}{c|}{30,000} &
  \multicolumn{4}{c|}{30,000} &
  \multicolumn{4}{c|}{25,000} &
  \multicolumn{4}{c|}{25.000} &
  \multicolumn{4}{c}{5,000} \\ \midrule
Conf Model &
  \multicolumn{2}{c}{LeNet-5} &
  \multicolumn{2}{c|}{CLIP ViT-B-16} &
  \multicolumn{2}{c}{LeNet-5} &
  \multicolumn{2}{c|}{CLIP ViT-B-16} &
  \multicolumn{2}{c}{WRN-28-2} &
  \multicolumn{2}{c|}{CLIP ViT-B-16} &
  \multicolumn{2}{c}{ResNet-18} &
  \multicolumn{2}{c|}{CLIP ViT-B-16} &
  \multicolumn{2}{c}{ResNet-18} &
  \multicolumn{2}{c}{CLIP ViT-B-16} \\ \midrule
Prior &
  0.5 &
  0.7 &
  0.5 &
  0.7 &
  0.5 &
  0.7 &
  0.5 &
  0.7 &
  0.4 &
  0.6 &
  0.4 &
  0.6 &
  0.4 &
  0.6 &
  0.4 &
  0.6 &
  0.4 &
  0.6 &
  0.4 &
  0.6 \\ \midrule
Sconf Abs &
  99.03\scriptsize{$\pm$0.01} &
  99.63\scriptsize{$\pm$0.09} &
  98.16\scriptsize{$\pm$0.11} &
  99.11\scriptsize{$\pm$0.29} &
  98.39\scriptsize{$\pm$0.24} &
  96.47\scriptsize{$\pm$0.11} &
  75.28\scriptsize{$\pm$0.91} &
  76.91\scriptsize{$\pm$1.04} &
  87.36\scriptsize{$\pm$1.22} &
  89.36\scriptsize{$\pm$0.82} &
  90.16\scriptsize{$\pm$1.32} &
  88.97\scriptsize{$\pm$0.12} &
  75.79\scriptsize{$\pm$0.27} &
  76.79\scriptsize{$\pm$0.21} &
  69.51\scriptsize{$\pm$0.44} &
  63.39\scriptsize{$\pm$0.30} &
  76.84\scriptsize{$\pm$0.75} &
  76.50\scriptsize{$\pm$0.68} &
  74.44\scriptsize{$\pm$0.78} &
  66.33\scriptsize{$\pm$1.87} \\
Sconf ReLU &
  99.45\scriptsize{$\pm$0.04} &
  99.65\scriptsize{$\pm$0.08} &
  98.02\scriptsize{$\pm$0.09} &
  \textbf{99.11\scriptsize{$\pm$0.30}} &
  98.23\scriptsize{$\pm$0.15} &
  96.40\scriptsize{$\pm$0.16} &
  76.14\scriptsize{$\pm$0.11} &
  76.91\scriptsize{$\pm$1.05} &
  88.56\scriptsize{$\pm$0.57} &
  89.66\scriptsize{$\pm$0.36} &
  90.50\scriptsize{$\pm$0.44} &
  88.79\scriptsize{$\pm$0.41} &
  74.95\scriptsize{$\pm$0.55} &
  76.83\scriptsize{$\pm$0.62} &
  69.67\scriptsize{$\pm$1.51} &
  63.77\scriptsize{$\pm$2.05} &
  77.40\scriptsize{$\pm$0.31} &
  76.51\scriptsize{$\pm$0.71} &
  75.26\scriptsize{$\pm$0.66} &
  67.12\scriptsize{$\pm$2.08} \\
Sconf NN Abs &
  99.63\scriptsize{$\pm$0.10} &
  99.48\scriptsize{$\pm$0.10} &
  98.51\scriptsize{$\pm$0.03} &
  99.19\scriptsize{$\pm$0.25} &
  98.41\scriptsize{$\pm$0.06} &
  96.32\scriptsize{$\pm$0.10} &
  76.42\scriptsize{$\pm$0.45} &
  77.57\scriptsize{$\pm$0.73} &
  89.04\scriptsize{$\pm$0.88} &
  88.97\scriptsize{$\pm$0.29} &
  89.05\scriptsize{$\pm$2.11} &
  87.76\scriptsize{$\pm$0.82} &
  74.55\scriptsize{$\pm$0.23} &
  75.82\scriptsize{$\pm$0.44} &
  68.93\scriptsize{$\pm$2.00} &
  63.79\scriptsize{$\pm$1.43} &
  77.55\scriptsize{$\pm$0.31} &
  75.97\scriptsize{$\pm$0.66} &
  75.66\scriptsize{$\pm$0.51} &
  65.80\scriptsize{$\pm$0.45} \\
Sconf Unbiased &
  99.15\scriptsize{$\pm$0.07} &
  99.64\scriptsize{$\pm$0.06} &
  98.44\scriptsize{$\pm$0.03} &
  99.16\scriptsize{$\pm$0.33} &
  98.05\scriptsize{$\pm$0.05} &
  95.98\scriptsize{$\pm$0.45} &
  76.38\scriptsize{$\pm$0.13} &
  77.37\scriptsize{$\pm$0.90} &
  88.72\scriptsize{$\pm$0.52} &
  87.90\scriptsize{$\pm$0.47} &
  88.71\scriptsize{$\pm$0.59} &
  88.86\scriptsize{$\pm$0.25} &
  72.87\scriptsize{$\pm$1.30} &
  73.23\scriptsize{$\pm$1.23} &
  69.55\scriptsize{$\pm$0.31} &
  64.51\scriptsize{$\pm$2.88} &
  77.76\scriptsize{$\pm$0.40} &
  76.74\scriptsize{$\pm$0.63} &
  74.36\scriptsize{$\pm$0.60} &
  67.42\scriptsize{$\pm$2.06} \\
\cellc \method &
\cellc  \textbf{99.90\scriptsize{$\pm$0.00}} &
\cellc  \textbf{99.89\scriptsize{$\pm$0.01}} &
\cellc  \textbf{98.67\scriptsize{$\pm$0.30}} &
\cellc  98.55\scriptsize{$\pm$0.08} &
\cellc  \textbf{98.58\scriptsize{$\pm$0.03}} &
\cellc  \textbf{98.48\scriptsize{$\pm$0.03}} &
\cellc  \textbf{76.78\scriptsize{$\pm$0.21}} &
\cellc  \textbf{78.47\scriptsize{$\pm$0.18}} &
\cellc  \textbf{95.97\scriptsize{$\pm$0.11}} &
\cellc  \textbf{95.93\scriptsize{$\pm$0.02}} &
\cellc  \textbf{97.88\scriptsize{$\pm$0.11}} &
\cellc  \textbf{97.60\scriptsize{$\pm$0.13}} &
\cellc  \textbf{85.58\scriptsize{$\pm$0.88}} &
\cellc  \textbf{87.85\scriptsize{$\pm$0.49}} &
\cellc  \textbf{87.94\scriptsize{$\pm$0.34}} &
\cellc  \textbf{86.56\scriptsize{$\pm$0.94}} &
\cellc  \textbf{78.64\scriptsize{$\pm$0.16}} &
\cellc  \textbf{78.33\scriptsize{$\pm$0.07}} &
\cellc  \textbf{79.06\scriptsize{$\pm$0.05}} &
\cellc  \textbf{78.69\scriptsize{$\pm$0.08}} \\ \bottomrule
\end{tabular}%
}
\end{table*}

\begin{table*}[t!]
\centering
\caption{Accuracy on confidence difference (ConfDiff) learning for binary classification. All results are averaged over three runs.}
\label{tab:appendix-confdiff}
\resizebox{\textwidth}{!}{%
\begin{tabular}{@{}l|cccc|cccc|cccc|cccc|cccc@{}}
\toprule
Dataset &
  \multicolumn{4}{c|}{F-MNIST} &
  \multicolumn{4}{c|}{MNIST} &
  \multicolumn{4}{c|}{CIFAR-10} &
  \multicolumn{4}{c|}{CIFAR-100} &
  \multicolumn{4}{c}{STL-10} \\ \midrule
\#Pairs &
  \multicolumn{4}{c|}{30,000} &
  \multicolumn{4}{c|}{30,000} &
  \multicolumn{4}{c|}{25,000} &
  \multicolumn{4}{c|}{25.000} &
  \multicolumn{4}{c}{5,000} \\ \midrule
Conf Model &
  \multicolumn{2}{c}{LeNet-5} &
  \multicolumn{2}{c|}{CLIP ViT-B-16} &
  \multicolumn{2}{c}{LeNet-5} &
  \multicolumn{2}{c|}{CLIP ViT-B-16} &
  \multicolumn{2}{c}{WRN-28-2} &
  \multicolumn{2}{c|}{CLIP ViT-B-16} &
  \multicolumn{2}{c}{ResNet-18} &
  \multicolumn{2}{c|}{CLIP ViT-B-16} &
  \multicolumn{2}{c}{ResNet-18} &
  \multicolumn{2}{c}{CLIP ViT-B-16} \\ \midrule
Prior &
  0.5 &
  0.7 &
  0.5 &
  0.7 &
  0.5 &
  0.7 &
  0.5 &
  0.7 &
  0.4 &
  0.6 &
  0.4 &
  0.6 &
  0.4 &
  0.6 &
  0.4 &
  0.6 &
  0.4 &
  0.6 &
  0.4 &
  0.6 \\ \midrule
ConfDiff Abs &
  99.82\scriptsize{$\pm$0.03} &
  99.36\scriptsize{$\pm$0.14}  &
  98.90\scriptsize{$\pm$0.36}  &
  93.80\scriptsize{$\pm$1.16}  &
  97.31\scriptsize{$\pm$0.08}  &
  92.11\scriptsize{$\pm$3.31}  &
  91.67\scriptsize{$\pm$0.91}  &
  89.19\scriptsize{$\pm$2.26}  &
  90.12\scriptsize{$\pm$4.19} &
  93.18\scriptsize{$\pm$0.42} &
  88.61\scriptsize{$\pm$7.50} &
  94.11\scriptsize{$\pm$0.29} &
  82.89\scriptsize{$\pm$0.32} &
  81.80\scriptsize{$\pm$1.95} &
  81.45\scriptsize{$\pm$0.26} &
  80.64\scriptsize{$\pm$1.33} &
  73.17\scriptsize{$\pm$2.06} &
  73.53\scriptsize{$\pm$3.19} &
  77.33\scriptsize{$\pm$0.74} &
  76.72\scriptsize{$\pm$0.57} \\
ConfDiff ReLU &
  99.83\scriptsize{$\pm$0.03} &
  99.31\scriptsize{$\pm$0.18} &
  99.29\scriptsize{$\pm$0.11} &
  95.54\scriptsize{$\pm$1.07} &
  97.42\scriptsize{$\pm$0.11} &
  93.28\scriptsize{$\pm$2.35} &
  90.26\scriptsize{$\pm$0.76} &
  88.97\scriptsize{$\pm$1.43} &
  90.36\scriptsize{$\pm$3.07} &
  93.05\scriptsize{$\pm$0.31} &
  88.78\scriptsize{$\pm$2.91} &
  93.86\scriptsize{$\pm$0.48} &
  83.13\scriptsize{$\pm$0.27} &
  82.64\scriptsize{$\pm$1.05} &
  81.68\scriptsize{$\pm$0.46} &
  80.75\scriptsize{$\pm$1.03} &
  72.39\scriptsize{$\pm$3.06} &
  73.41\scriptsize{$\pm$1.83} &
  77.59\scriptsize{$\pm$0.17} &
  76.11\scriptsize{$\pm$1.54} \\
ConfDiff Unbiased &
  99.83\scriptsize{$\pm$0.04} &
  99.31\scriptsize{$\pm$0.21} &
  99.43\scriptsize{$\pm$0.10} &
  97.59\scriptsize{$\pm$0.27} &
  97.34\scriptsize{$\pm$0.09} &
  94.46\scriptsize{$\pm$1.86} &
  89.22\scriptsize{$\pm$1.39} &
  88.05\scriptsize{$\pm$0.68} &
  90.05\scriptsize{$\pm$5.23} &
  93.23\scriptsize{$\pm$0.33} &
  87.91\scriptsize{$\pm$9.03} &
  93.87\scriptsize{$\pm$0.38} &
  83.65\scriptsize{$\pm$0.11} &
  82.69\scriptsize{$\pm$1.08} &
  81.94\scriptsize{$\pm$0.43} &
  80.89\scriptsize{$\pm$0.88} &
  72.13\scriptsize{$\pm$2.70} &
  73.53\scriptsize{$\pm$3.19} &
  77.98\scriptsize{$\pm$0.08} &
  77.81\scriptsize{$\pm$0.40} \\
\cellc \method &
\cellc  \textbf{99.88\scriptsize{$\pm$0.01}} &
\cellc  \textbf{98.43\scriptsize{$\pm$0.33}} &
\cellc  \textbf{99.86\scriptsize{$\pm$0.01}} &
\cellc  \textbf{99.42\scriptsize{$\pm$0.09}} &
\cellc  \textbf{98.30\scriptsize{$\pm$0.05}} &
\cellc  \textbf{97.74\scriptsize{$\pm$0.07}} &
\cellc  \textbf{93.41\scriptsize{$\pm$0.12}} &
\cellc  \textbf{91.88\scriptsize{$\pm$0.08}} &
\cellc  \textbf{95.36\scriptsize{$\pm$0.19}} &
\cellc  \textbf{94.81\scriptsize{$\pm$0.03}} &
\cellc  \textbf{96.14\scriptsize{$\pm$0.67}} &
\cellc  \textbf{96.23\scriptsize{$\pm$0.11}} &
\cellc  \textbf{86.12\scriptsize{$\pm$0.76}} &
\cellc  \textbf{84.95\scriptsize{$\pm$1.20}} &
\cellc  \textbf{83.42\scriptsize{$\pm$1.12}} &
\cellc  \textbf{82.53\scriptsize{$\pm$0.27}} &
\cellc  \textbf{77.99\scriptsize{$\pm$0.75}} &
\cellc  \textbf{76.24\scriptsize{$\pm$0.42}} &
\cellc  \textbf{78.49\scriptsize{$\pm$0.31}} &
\cellc  \textbf{78.57\scriptsize{$\pm$0.28}} \\ \bottomrule
\end{tabular}%
}
\end{table*}

\subsection{Unlabeled Data}

\subsubsection{Setup}
\label{sec:appendix-exp-ulb-setup}

\begin{table}[h!]
\centering
\caption{Hyper-parameters for positive unlabeled (PosUlb), unlabeled unlabeled (UlbUlb), and similarity dsimilarity unlabeled (SimDsimUlb) learning used in experiments.}
\label{tab:append-param-ulb}
\resizebox{0.55 \textwidth}{!}{%
\begin{tabular}{@{}c|cccc@{}}
\toprule
\multicolumn{1}{c|}{Hyper-parameter}   & MNIST \& F-MNIST & CIFAR-10 & CIFAR-100 & STL-10 \\ \midrule
Image Size & 28 & 32 & 32 & 96  \\
Model      &  LeNet-5 & WRN-28-2 &   ResNet-18        &    ResNet-18   \\
Batch Size        &     64     &    64       &   64 &  32     \\
Optimizer & AdamW & AdamW & AdamW & AdamW \\
Learning Rate     &     5e-4     &   1e-3        &  1e-3 &  1e-3     \\
Weight Decay      &     1e-4     &   1e-3       &   1e-3    & 1e-3  \\
LR Scheduler          &     Cosine     &   Cosine   &   Cosine  & Cosine     \\
Training Epochs   &     50    &   50        &  50 & 50      \\
\bottomrule
\end{tabular}%
}
\end{table}

For unlabeled data, we evaluate on PosUlb, UlbUlb, and SDUlb settings with class priors. 
The hyper-parameters are shown in \cref{tab:append-param-ulb}. 
For PosUlb, we sample labeled set from only positive samples, and form the unlabeled set with both positive and negative samples whose distribution follows the class prior. 
For UlbUlb, we form both unlabeled set similarly as in PosUlb. 
For SDUlb, the labeled pairwise observation is formed similarly as in PSim.

\subsubsection{Results}
\label{sec:appendix-exp-ulb-results}

We present more results of PosUlb in \cref{tab:appendix-pu}, and evaluation on UlbUlb and SDUlb in \cref{tab:appendix-uulearn} and \cref{tab:appendix-SDUlb} respectively.
Our approach achieves the best results across different settings except on F-MNIST of UlbUlb evaluation.

\begin{table*}[t!]
\centering
\caption{Accuracy on positive unlabeled (PU) learning for binary classification. All results are averaged over three runs.}
\label{tab:appendix-pu}
\resizebox{\textwidth}{!}{%
\begin{tabular}{@{}l|ccc|ccc|ccc|ccc|ccc@{}}
\toprule
 &
  \multicolumn{3}{c|}{FMNIST} &
  \multicolumn{3}{c|}{MNIST} &
  \multicolumn{3}{c|}{CIFAR-10} &
  \multicolumn{3}{c|}{CIFAR-100} &
  \multicolumn{3}{c}{STL-10} \\ \midrule
 Prior &
  \multicolumn{3}{c|}{0.3} &
  \multicolumn{3}{c|}{0.5} &
  \multicolumn{3}{c|}{0.4} &
  \multicolumn{3}{c|}{0.4} &
  \multicolumn{3}{c}{0.4} \\ \midrule
\# Pos &
  100 &
  500 &
  1000 &
  100 &
  500 &
  1000 &
  500 &
  1000 &
  2000 &
  1000 &
  2000 &
  4000 &
  500 &
  1000 &
  2000 \\ \midrule
Count Loss &
  99.48\scriptsize{$\pm$0.25} &
  99.72\scriptsize{$\pm$0.01} &
  99.77\scriptsize{$\pm$0.02} &
  85.25\scriptsize{$\pm$0.53} &
  93.27\scriptsize{$\pm$0.65} &
  95.14\scriptsize{$\pm$0.49} &
  87.76\scriptsize{$\pm$0.59} &
  88.61\scriptsize{$\pm$0.68} &
  87.96\scriptsize{$\pm$0.42} &
  70.57\scriptsize{$\pm$1.50} &
  78.13\scriptsize{$\pm$0.19} &
  78.17\scriptsize{$\pm$2.49} &
  77.11\scriptsize{$\pm$0.60} &
  78.79\scriptsize{$\pm$0.96} &
  79.77\scriptsize{$\pm$1.40} \\
CVIR &
  97.77\scriptsize{$\pm$0.98} &
  99.24\scriptsize{$\pm$0.22} &
  99.31\scriptsize{$\pm$0.37} &
  74.67\scriptsize{$\pm$1.83} &
  93.61\scriptsize{$\pm$1.08} &
  95.95\scriptsize{$\pm$0.26} &
  88.65\scriptsize{$\pm$2.59} &
  93.37\scriptsize{$\pm$0.24} &
  94.69\scriptsize{$\pm$0.24} &
  78.56\scriptsize{$\pm$0.22} &
  82.94\scriptsize{$\pm$0.37} &
  85.22\scriptsize{$\pm$1.07} &
  77.67\scriptsize{$\pm$1.11} &
  81.84\scriptsize{$\pm$1.10} &
  85.38\scriptsize{$\pm$0.83} \\
Dist PU &
  96.87\scriptsize{$\pm$0.24} &
  97.23\scriptsize{$\pm$0.31} &
  97.44\scriptsize{$\pm$0.15} &
  85.60\scriptsize{$\pm$1.69} &
  90.94\scriptsize{$\pm$1.87} &
  95.36\scriptsize{$\pm$0.38} &
  83.61\scriptsize{$\pm$4.52} &
  82.60\scriptsize{$\pm$2.48} &
  85.38\scriptsize{$\pm$1.13} &
  69.12\scriptsize{$\pm$1.39} &
  69.83\scriptsize{$\pm$1.43} &
  70.82\scriptsize{$\pm$1.12} &
  71.07\scriptsize{$\pm$1.12} &
  70.89\scriptsize{$\pm$0.63} &
  69.86\scriptsize{$\pm$1.07} \\
NN PU &
  95.27\scriptsize{$\pm$1.40} &
  98.63\scriptsize{$\pm$0.54} &
  99.18\scriptsize{$\pm$0.10} &
  67.99\scriptsize{$\pm$0.80} &
  77.84\scriptsize{$\pm$0.35} &
  78.77\scriptsize{$\pm$1.50} &
  87.45\scriptsize{$\pm$0.66} &
  90.32\scriptsize{$\pm$0.50} &
  86.60\scriptsize{$\pm$1.26} &
  75.49\scriptsize{$\pm$0.88} &
  77.26\scriptsize{$\pm$0.47} &
  78.53\scriptsize{$\pm$0.95} &
  74.57\scriptsize{$\pm$0.54} &
  77.32\scriptsize{$\pm$0.95} &
  77.68\scriptsize{$\pm$2.07} \\
U PU &
  85.09\scriptsize{$\pm$0.53} &
  84.90\scriptsize{$\pm$0.26} &
  84.85\scriptsize{$\pm$0.43} &
  74.56\scriptsize{$\pm$0.79} &
  79.58\scriptsize{$\pm$1.27} &
  79.41\scriptsize{$\pm$0.16} &
  81.38\scriptsize{$\pm$2.17} &
  87.51\scriptsize{$\pm$0.24} &
  90.54\scriptsize{$\pm$0.11} &
  68.70\scriptsize{$\pm$0.79} &
  70.08\scriptsize{$\pm$0.98} &
  74.07\scriptsize{$\pm$1.51} &
  73.37\scriptsize{$\pm$0.57} &
  75.31\scriptsize{$\pm$0.74} &
  79.08\scriptsize{$\pm$0.76} \\
Var PU &
  95.02\scriptsize{$\pm$0.02} &
  97.52\scriptsize{$\pm$1.48} &
  99.22\scriptsize{$\pm$0.26} &
  48.61\scriptsize{$\pm$0.64} &
  48.61\scriptsize{$\pm$0.01} &
  51.22\scriptsize{$\pm$2.88} &
  77.00\scriptsize{$\pm$2.82} &
  84.45\scriptsize{$\pm$2.58} &
  87.34\scriptsize{$\pm$1.80} &
  61.02\scriptsize{$\pm$0.22} &
  66.02\scriptsize{$\pm$0.29} &
  70.57\scriptsize{$\pm$1.94} &
  60.98\scriptsize{$\pm$0.78} &
  62.37\scriptsize{$\pm$1.44} &
  62.18\scriptsize{$\pm$1.02} \\
\cellc \method &
\cellc  \textbf{99.55\scriptsize{$\pm$0.21}} &
\cellc  \textbf{99.84\scriptsize{$\pm$0.03}} &
\cellc  \textbf{99.87\scriptsize{$\pm$0.01}} &
\cellc  \textbf{87.57\scriptsize{$\pm$0.21}} &
\cellc  \textbf{95.04\scriptsize{$\pm$0.74}} &
\cellc  \textbf{96.85\scriptsize{$\pm$0.17}} &
\cellc  \textbf{91.67\scriptsize{$\pm$0.19}} &
\cellc  \textbf{93.69\scriptsize{$\pm$0.28}} &
\cellc  \textbf{94.80\scriptsize{$\pm$0.12}} &
\cellc  \textbf{80.30\scriptsize{$\pm$0.12}} &
\cellc  \textbf{83.32\scriptsize{$\pm$0.23}} &
\cellc  \textbf{85.98\scriptsize{$\pm$0.29}} &
\cellc  \textbf{79.60\scriptsize{$\pm$0.95}} &
\cellc  \textbf{82.87\scriptsize{$\pm$0.83}} &
\cellc  \textbf{87.51\scriptsize{$\pm$0.60}} \\ \bottomrule
\end{tabular}%
}
\end{table*}

\begin{table*}[t!]
\centering
\caption{Accuracy on unlabeled unlabeled (UU) learning for binary classification. All results are averaged over three runs.}
\label{tab:appendix-uulearn}
\resizebox{\textwidth}{!}{%
\begin{tabular}{@{}l|ccc|ccc|ccc|ccc|ccc@{}}
\toprule
 &
  \multicolumn{3}{c|}{FMNIST} &
  \multicolumn{3}{c|}{MNIST} &
  \multicolumn{3}{c|}{CIFAR-10} &
  \multicolumn{3}{c|}{CIFAR-100} &
  \multicolumn{3}{c}{STL-10} \\ \midrule
\# Ulb &
  10,000 &
  30,000 &
  30,000 &
  10,000 &
  30,000 &
  30,000 &
  10,000 &
  25,000 &
  25,000 &
  10,000 &
  25,000 &
  25,000 &
  2,500 &
  5,000 &
  5,000 \\
(Prior1, Prior2) &
  (0.4, 0.6) &
  (0.4, 0.6) &
  (0.8, 0.2) &
  (0.4, 0.6) &
  (0.4, 0.6) &
  (0.8, 0.2) &
  (0.4, 0.6) &
  (0.4, 0.6) &
  (0.8, 0.2) &
  (0.4, 0.6) &
  (0.4, 0.6) &
  (0.8, 0.2) &
  (0.4, 0.6) &
  (0.4, 0.6) &
  (0.8, 0.2) \\ \midrule
UU Learn &
  \textbf{99.08\scriptsize{$\pm$0.21}} &
  \textbf{99.54\scriptsize{$\pm$0.05}} &
  \textbf{99.89\scriptsize{$\pm$0.02}} &
  89.75\scriptsize{$\pm$0.41} &
  93.24\scriptsize{$\pm$1.02} &
  98.19\scriptsize{$\pm$0.10} &
  88.32\scriptsize{$\pm$1.64} &
  89.42\scriptsize{$\pm$0.11} &
  94.46\scriptsize{$\pm$0.05} &
  66.23\scriptsize{$\pm$1.59} &
  66.37\scriptsize{$\pm$3.05} &
  71.23\scriptsize{$\pm$0.54} &
  78.59\scriptsize{$\pm$1.97} &
  72.55\scriptsize{$\pm$1.27} &
  81.42\scriptsize{$\pm$1.31} \\
\cellc \method &
\cellc  97.65\scriptsize{$\pm$0.75} &
\cellc  96.10\scriptsize{$\pm$0.75} &
\cellc  99.87\scriptsize{$\pm$0.01} &
\cellc  \textbf{97.27\scriptsize{$\pm$0.24}} &
\cellc  \textbf{96.87\scriptsize{$\pm$0.80}} &
\cellc  \textbf{98.91\scriptsize{$\pm$0.04}} &
\cellc  \textbf{94.45\scriptsize{$\pm$0.45}} &
\cellc  \textbf{95.01\scriptsize{$\pm$0.44}} &
\cellc  \textbf{98.07\scriptsize{$\pm$0.07}} &
\cellc  \textbf{74.66\scriptsize{$\pm$1.29}} &
\cellc  \textbf{76.75\scriptsize{$\pm$3.05}} &
\cellc  \textbf{89.83\scriptsize{$\pm$0.44}} &
\cellc  \textbf{82.09\scriptsize{$\pm$2.33}} &
\cellc  \textbf{84.78\scriptsize{$\pm$1.19}} &
\cellc  \textbf{88.27\scriptsize{$\pm$0.96}} \\ \bottomrule
\end{tabular}%
}
\end{table*}

\begin{table*}[t!]
\centering
\caption{Accuracy on similarity dissimilarity unlabeled (SDUlb) learning for binary classification. All results are averaged over three runs.}
\label{tab:appendix-simdsimulb}
\resizebox{\textwidth}{!}{%
\begin{tabular}{@{}l|ccc|ccc|ccc|ccc|ccc@{}}
\toprule
             & \multicolumn{3}{c|}{FMNIST} & \multicolumn{3}{c|}{MNIST} & \multicolumn{3}{c|}{CIFAR-10} & \multicolumn{3}{c|}{CIFAR-100} & \multicolumn{3}{c}{STL-10} \\ \midrule
\# Sim Pair  & 0       & 5,000   & 10,000  & 0       & 5,000   & 10,000 & 0        & 5,000    & 10,000  & 0        & 5,000    & 10,000   & 0       & 1,000   & 2,000  \\
\# Dsim Pair & 10,000  & 5,000   & 0       & 10,000  & 5,000   & 0      & 10,000   & 5,000    & 0       & 10,000   & 5,000    & 0        & 2,000   & 1,000   & 0      \\
\# Ulb       & 20,000  & 20,000  & 20,000  & 20,000  & 20,000  & 20,000 & 20,000   & 20,000   & 20,000  & 20,000   & 20,000   & 20,000   & 4,000   & 4,000   & 4,000  \\ \midrule
RiskSD &
  87.61\scriptsize{$\pm$0.48} &
  89.32\scriptsize{$\pm$0.15} &
  87.06\scriptsize{$\pm$0.97} &
  87.18\scriptsize{$\pm$0.10} &
  89.09\scriptsize{$\pm$1.89} &
  83.02\scriptsize{$\pm$1.24} &
  78.69\scriptsize{$\pm$3.56} &
  84.61\scriptsize{$\pm$0.50} &
  79.09\scriptsize{$\pm$2.36} &
  65.36\scriptsize{$\pm$0.28} &
  65.87\scriptsize{$\pm$1.23} &
  63.43\scriptsize{$\pm$1.45} &
  66.94\scriptsize{$\pm$1.42} &
  66.92\scriptsize{$\pm$2.86} &
   65.02\scriptsize{$\pm$1.59} \\
\cellc \method &
\cellc  \textbf{92.81\scriptsize{$\pm$0.08}} &
\cellc  \textbf{92.57\scriptsize{$\pm$0.24}} &
\cellc  \textbf{92.24\scriptsize{$\pm$1.20}} &
\cellc  \textbf{97.27\scriptsize{$\pm$0.11}} &
\cellc  \textbf{96.54\scriptsize{$\pm$0.36}} &
\cellc  \textbf{96.41\scriptsize{$\pm$0.95}} &
\cellc  \textbf{93.12\scriptsize{$\pm$0.28}} &
\cellc  \textbf{92.76\scriptsize{$\pm$0.82}} &
\cellc  \textbf{84.24\scriptsize{$\pm$1.20}} &
\cellc  \textbf{78.59\scriptsize{$\pm$1.32}} &
\cellc  \textbf{73.02\scriptsize{$\pm$1.54}} &
\cellc  \textbf{70.24\scriptsize{$\pm$1.16}} &
\cellc  \textbf{79.28\scriptsize{$\pm$0.98}} &
\cellc  \textbf{75.93\scriptsize{$\pm$1.69}} &
\cellc  \textbf{75.67\scriptsize{$\pm$1.26}} \\ \bottomrule
\end{tabular}%
}
\end{table*}

\subsection{Other Settings}
\label{sec:appendix-exp-other}

Here we present the evaluation of other weak supervision settings. 

\subsubsection{Positive Confidence Learning}

We evaluation on positive confidence (PosConf) learning \cite{ishida2018binary}, where the weak supervision is given as the confidence score of a sample being positive, from the pre-trained models.
The NFA of PosConf consists of $L$ states for each instances in the training batch, and allows transition via both 0 and 1. Each positive transition path is weighted by the positive confidence score $c$, and each negative transition path is weighted by $1 -c$. 
This modeling can also be easily extended to subset confidence learning \cite{cao2021multi} and soft label learning \cite{ishida2022performance}.

The training parameters follow \cref{tab:append-param-pair} and the results are shown in \cref{tab:appendix-posconf}.
Our method outperforms the baseline PConf \cite{ishida2018binary} except on MNIST.

\begin{table}[h!]
\centering
\caption{Accuracy on positive confidence (PosConf) learning for binary classification. All results are averaged over three runs.}
\resizebox{0.9\textwidth}{!}{%
\begin{tabular}{@{}c|ccc|ccc|ccc|ccc@{}}
\toprule
Dataset    & \multicolumn{3}{c|}{MNIST}        & \multicolumn{3}{c|}{F-MNIST}      & \multicolumn{3}{c|}{CIFAR-10}       & \multicolumn{3}{c}{CIFAR-100}         \\ \midrule
\# Data    & 15,000  & 30,000  & 30,000        & 15,000  & 30,000  & 30,000        & 10,000   & 25,000   & 25,000        & 10,000    & 25,000    & 25,000        \\ \midrule
Conf Model & LeNet-5 & LeNet-5 & CLIP ViT-B-16 & LeNet-5 & LeNet-5 & CLIP ViT-B-16 & WRN-28-2 & WRN-28-2 & CLIP ViT-B-16 & ResNet-18 & ResNet-18 & CLIP ViT-B-16 \\ \midrule
PConf &
  \textbf{79.61\scriptsize{$\pm$0.65}} &
  \textbf{80.55\scriptsize{$\pm$0.84}} &
  \textbf{80.23\scriptsize{$\pm$0.84}} &
  91.25\scriptsize{$\pm$0.23} &
  90.67\scriptsize{$\pm$0.13} &
  91.10\scriptsize{$\pm$0.13} &
  90.71\scriptsize{$\pm$1.86} &
  92.74\scriptsize{$\pm$0.26} &
  92.70\scriptsize{$\pm$0.22} &
  74.09\scriptsize{$\pm$1.61} &
  79.39\scriptsize{$\pm$0.50} &
  79.49\scriptsize{$\pm$0.62} \\
\cellc \method &
 \cellc 79.52\scriptsize{$\pm$0.56} &
\cellc  80.09\scriptsize{$\pm$0.36} &
\cellc  79.89\scriptsize{$\pm$0.56} &
\cellc  \textbf{92.22\scriptsize{$\pm$0.06}} &
\cellc  \textbf{91.95\scriptsize{$\pm$0.16}} &
\cellc  \textbf{92.05\scriptsize{$\pm$0.09}} &
\cellc  \textbf{95.31\scriptsize{$\pm$0.13}} &
\cellc  \textbf{96.84\scriptsize{$\pm$0.12}} &
\cellc  \textbf{96.93\scriptsize{$\pm$0.14}} &
\cellc  \textbf{83.70\scriptsize{$\pm$0.13}} &
\cellc  \textbf{86.62\scriptsize{$\pm$0.18}} &
\cellc  \textbf{87.24\scriptsize{$\pm$0.13}} \\ \bottomrule
\end{tabular}%
}
\label{tab:appendix-posconf}
\end{table}


\end{document}